\documentclass[letterpaper]{article} 

\usepackage{aaai24}  
\usepackage{times}  
\usepackage{helvet}  
\usepackage{courier}  
\usepackage[hyphens]{url}  
\usepackage{graphicx} 
\usepackage{natbib}  
\usepackage{caption} 
\usepackage{algorithm}
\usepackage{algorithmic}
\usepackage{multirow} 
\usepackage{booktabs}

\usepackage{newfloat}
\usepackage{listings}
\DeclareCaptionStyle{ruled}{labelfont=normalfont,labelsep=colon,strut=off} 
\floatstyle{ruled}
\newfloat{listing}{tb}{lst}{}
\floatname{listing}{Listing}
\usepackage{algorithm}
\usepackage{algorithmic}
\usepackage{pifont}
\usepackage{xcolor,colortbl}
\usepackage{amsmath, amsfonts, amssymb}
\usepackage{bm}

\usepackage{xspace}
\makeatletter
\DeclareRobustCommand\onedot{\futurelet\@let@token\@onedot}
\def\@onedot{\ifx\@let@token.\else.\null\fi\xspace}
\DeclareMathOperator*{\fe}{\mathbb{E}}
\def\eg{\emph{e.g}\onedot} 
\def\ie{\emph{i.e}\onedot} 
\def\cf{\emph{c.f}\onedot} 
 
\def\wrt{w.r.t\onedot} 
\def\etal{\emph{et al}\onedot}
\makeatother

\newcommand{\domw}{\mathcal{W}}
\newcommand{\domx}{\mathcal{X}}
\newcommand{\domy}{\mathcal{Y}}
\newcommand{\cmark}{\ding{51}}%
\newcommand{\xmark}{\ding{55}}%

\definecolor{gray0}{gray}{0.95}
\definecolor{gray1}{gray}{0.85}
\definecolor{gray2}{gray}{0.75}
\newcommand\blankfootnote[1]{%
  \let\thefootnote\relax\footnotetext{#1}%
  \let\thefootnote\svthefootnote%
}

\title{Federated Learning via Input-Output Collaborative Distillation}
\author{
    Xuan Gong\textsuperscript{\rm 1,\rm 3 \equalcontrib},
    Shanglin Li\textsuperscript{\rm 2 \equalcontrib},
    Yuxiang Bao\textsuperscript{\rm 2 \equalcontrib},
    Barry Yao\textsuperscript{\rm 1,\rm 4},
    Yawen Huang\textsuperscript{\rm 5},
    Ziyan Wu\textsuperscript{\rm 6},
    Baochang Zhang\textsuperscript{\rm 2,\rm 7,\rm 8,\rm 9 \dag},
    Yefeng Zheng\textsuperscript{\rm 5},
    David Doermann\textsuperscript{\rm 1 \dag}
}
\affiliations{
    \textsuperscript{\rm 1} University at Buffalo, Buffalo, NY, USA 
    \textsuperscript{\rm 2} Institute of Artificial Intelligence, Beihang University, Beijing, China \\
    \textsuperscript{\rm 3} Harvard Medical School, Boston, MA, USA 
    \textsuperscript{\rm 4} Virginia Tech, Blacksburg, VA, USA  \\
    \textsuperscript{\rm 5} Jarvis Research Center, Tencent YouTu Lab, Shenzhen, China
    \textsuperscript{\rm 6} United Imaging Intelligence, Burlington, MA, USA \\
    \textsuperscript{\rm 7} Hangzhou Research Institute, Beihang University, Hangzhou, China
    \textsuperscript{\rm 8} Zhongguancun Laboratory, Beijing, China \\
    \textsuperscript{\rm 9} Nanchang Institute of Technology, Nanchang, China

    xuangong@buffalo.edu, shanglin@buaa.edu.cn, yxbao@buaa.edu.cn, barryyao@vt.edu, yawenhuang@tencent.com, ziyan.wu@uii-ai.com, bczhang@buaa.edu.cn, yefengzheng@tencent.com, doermann@buffalo.edu    

}

\begin{document}
\maketitle

\begin{abstract}
   Federated learning (FL) is a machine learning paradigm in which distributed local nodes collaboratively train a central model without sharing individually held private data. Existing FL methods either iteratively share local model parameters or deploy co-distillation. However, the former is highly susceptible to private data leakage, and the latter design relies on the prerequisites of task-relevant real data. Instead, we propose a data-free FL framework based on local-to-central collaborative distillation with direct input and output space exploitation. Our design eliminates any requirement of recursive local parameter exchange or auxiliary task-relevant data to transfer knowledge, thereby giving direct privacy control to local users. In particular, to cope with the inherent data heterogeneity across locals, our technique learns to distill input on which each local model produces consensual yet unique results to represent each expertise. Our proposed FL framework achieves notable privacy-utility trade-offs with extensive experiments on image classification and segmentation tasks under various real-world heterogeneous federated learning settings on both natural and medical images. Code is available at  \url{https://github.com/lsl001006/FedIOD}. 
\end{abstract}
\renewcommand{\thefootnote}{\fnsymbol{footnote}}

\begin{figure}[h]
\centering
\includegraphics[width=\linewidth]{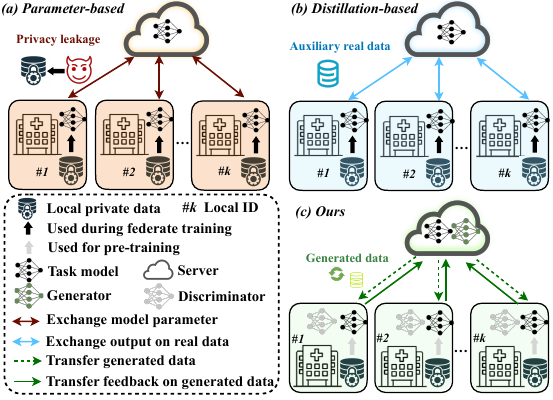}
\caption{ (a) Parameter-based methods recursively exchange model parameters between each local and server-side \cite{mcmahan2017communication,li2018federated, karimireddy2019scaffold}, which is highly vulnerable to a security attack \cite{zhu2019deep}. 
(b) Distillation-based methods utilize auxiliary task-dependent real data to conduct co-distillation between each local and the central server \cite{li2019fedmd, gong2022preserving}. 
(c) Our FL method conducts one-way distillation from locals to the server with generated data, eliminating the prerequisite of additional data required by typical distillation, and the security weaknesses of white-box attacks caused by recursive parameter exchange.
} 
\label{fig1}
\end{figure}

\section{Introduction}
\label{sec:intro}
The recent success of deep learning in various applications can be attributed to data-driven algorithms typically trained in a centralized fashion, \ie, computational units and data samples residing on the same server. Real-world scenarios, however, tend to disperse this wealth of data throughout separate locations and governed by diverse entities. Due to privacy regulations and communication limitations, collecting all data in one location for centralized training is often impractical, especially true for mobile vision and medical applications.

Accordingly, federated learning (FL) does not necessarily need all data samples to be centralized; instead, it relies on model fusion/distillation techniques to train one centralized model in a decentralized fashion. Privacy is a critical consideration, and it is vital to prevent private data leakage. Another challenge is data heterogeneity among locals, as distributed data centers tend to collect data in different settings. 

Most federated learning methods are based on the recursive exchange of local model parameters during the training process~\cite{mcmahan2017communication, li2018federated, karimireddy2019scaffold}. Each local node uploads its model parameters after a particular time of local training. The central server aggregates the parameters of the local model with different schemes~\cite{wang2020federated, li2019fair, hsu2020federated} and then distributes the aggregated parameters. Each local node receives the latest parameters to update its local model accordingly and continues with the next round of local training. 
However, naively employing such iterative parameter exchange suffers from known weaknesses: (1) All participating models must have exactly homogeneous architectures. (2) Iteratively sharing the model parameters opens all internal states of the model to white-box inference attacks, resulting in significant privacy leakage~\cite{chang2019cronus}. Recent works \cite{zhu2019deep, geiping2020inverting} obtain private training data from publicly shared model gradients.

Distillation-based methods are proposed to train the central model with aggregated locally-computed logits~\cite{li2019fedmd, lin2020ensemble, gong2022preserving}, eliminating the requirement of identical network architectures. 
However, to transfer knowledge, additional public data are commonly assumed to be accessible and sampled from the same underlying distribution as the privately held local data. 
This assumption can be strong in practice and unavoidably exposes private data to stealthy attacks. 
Although \cite{zhu2021data, zhang2022fedzkt, zhang2022fine} takes a step further to eliminate the requirement of real data for distillation, iterative model parameter exchange is still essential in these frameworks where knowledge transfer is only an auxiliary module for fine-tuning. As noted above, such parameter exchange is limited by identical model architecture and, more importantly, highly susceptible to privacy leakage. 
These methods require such recursive parameter exchange primarily because they mainly focus on the output distillation, leaving the input space under-explored. 

\begin{figure*}
\centering
\includegraphics[width=\linewidth]{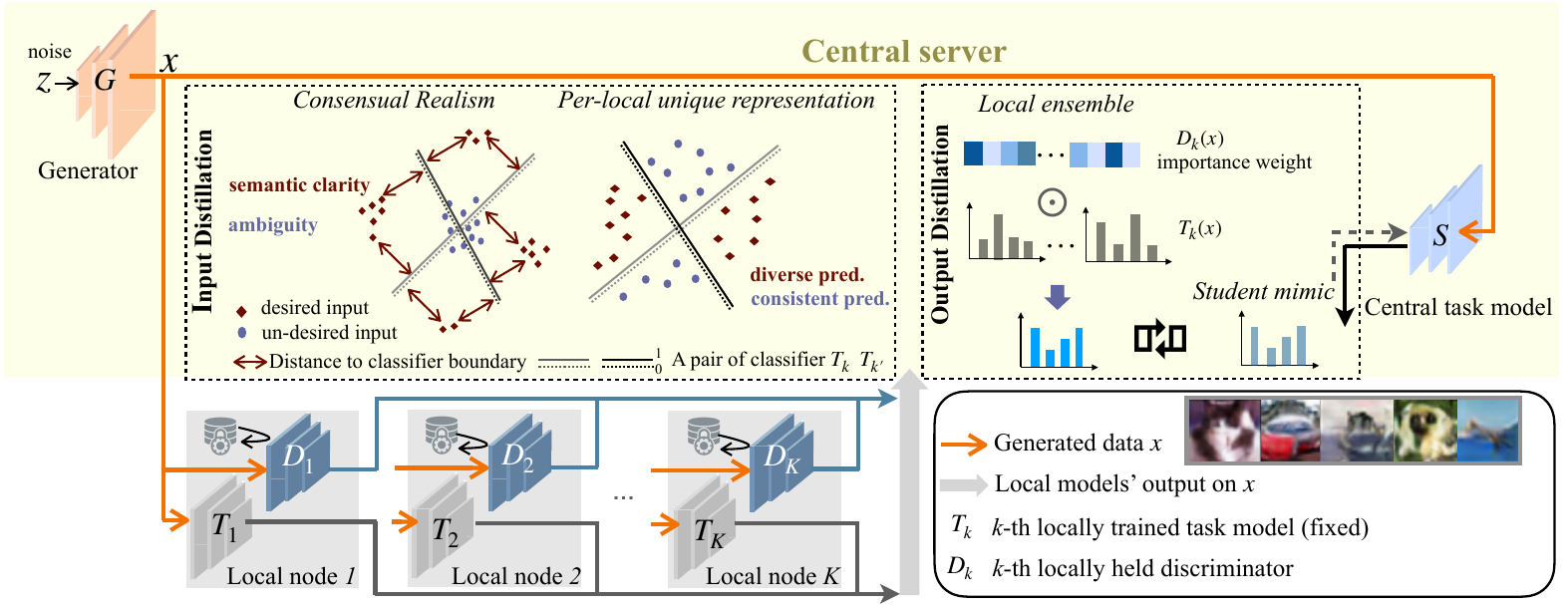}
\caption{\textbf{The overall pipeline of the proposed FedIOD.} We conduct distillation in input and output spaces to transfer knowledge from the locally trained task model $T_k$ and the auxiliary discriminator $D_k$ to the central task model $S$. \textbf{Input distillation} optimizes central generator $G$ to generate transferred input on which local models (1) achieve consensus on its semantic clarity, 
(2) and simultaneously produce diverse predictions. The latter is to exploit each local's unique expertise under the heterogeneous FL setting. 
\textbf{Output distillation} leverages per-input importance for output ensemble knowledge transfer. 
} 
\label{fig2}
\end{figure*}

In this paper, we propose a new federated learning framework  (FedIOD) that conducts a collaborative knowledge distillation in both the input and output space (as Figure \ref{fig1}).
It is purely based on data-free distillation without any prerequisite of auxiliary real data or locally trained model parameters. Besides, we adopt differential privacy protection on the gradients used to train the generator \cite{torkzadehmahani2019dp, chen2020gs}.
This, by design, gives explicit privacy control to each local node. 
Unlike the previous data-free federated distillation counterparts~\cite{zhu2021data, zhang2022fedzkt, zhang2022fine}, which employ both bidirectional distillation and iterative model parameter exchange, our framework makes another difference by conducting one-way distillation from thoroughly trained local models to the central model. These fully trained teacher models immediately enable us to explore the input space and learn the most efficient samples for knowledge distillation. Our critical insight is that each local's unique expertise under the heterogeneous FL setting can be further exploited. Therefore, we implement the input distillation according to the corresponding local products (\cf, Figure \ref{fig2}). This involves learning the transferred input to enable local nodes to reach a consensus on its semantic clarity while simultaneously generating diverse predictions with each task model. The former ensures the fundamental viability of the input data for transferring knowledge. At the same time, the latter allows the input data to leverage the unique aspects of each local node under heterogeneous federated learning scenarios.
Such feedback from local nodes enables us to deploy per-input importance weight for output ensemble distillation. 
We demonstrate the effectiveness of our proposed method on natural and medical images through comprehensive experiments on image classification and segmentation tasks under various real-world federated learning scenarios, including the most challenging cross-domain cross-site settings. 
Our key contributions can be summarized as follows. 
 \begin{itemize}
     \item
We propose a federated learning framework with collaborative distillation in both the input and output space. It eliminates any requirement on model parameter exchange, model structure identity, prior knowledge of the local task, or auxiliary real data. 
     \item
 To cope with the inherent heterogeneity of decentralized clients in federated learning, we introduce an ensemble distillation scheme that learns transferred input with explicit exploitation of each local's consensual and unique expertise. 
     \item
We conduct extensive experiments with natural and medical images on classification and segmentation tasks, demonstrating state-of-the-art privacy-utility trade-offs compared to the prior art.
\end{itemize}

\section{Related Work}
\label{sec:related}
\subsection{Knowledge Distillation}

Hinton \etal \cite{hinton2015distilling} first proposed the concept of knowledge distillation \ie, using a cumbersome network as a teacher to generate soft labels to supervise the training of a compact student network. 
Although most of the following works transfer knowledge with one teacher, some techniques focus on multiple teachers and propose a variety of aggregation schemes, \eg, gate learning in the supervised setting \cite{asif2019ensemble, xiang2020learning}, and relative sample similarity for unsupervised scenarios \cite{wu2019distilled}. Recent progress in data-free knowledge transfer \cite{fang2019data, chen2019data} focuses on an adversarial training scheme to generate hard-to-learn and hard-to-mimic samples. Similarly, DeepInversion \cite{yin2020dreaming} utilizes backpropagated gradients to generate transfer samples that cause disagreements between the teacher and the student. \cite{nayak2019zero} crafts a transfer set by modeling and fitting data distributions in output similarities. 

\subsection{Distillation-based Federated Learning}

Beyond the parameter based FL \cite{mcmahan2017communication, hsu2019measuring,li2018federated}, early FL works like \cite{jeong2018communication} employ parameter and model output exchanges. Although the following works \cite{li2019fedmd, chang2019cronus, li2021practical} are purely based on the output of the local model for knowledge transfer, the selection of transfer data is highly dependent on prior knowledge of private data (\ie, they are under similar data distributions). Some recently proposed methods \cite{lin2020ensemble, gong2022preserving} provide some relaxation on transfer data. However, it is still necessary to carefully select the transfer data according to prior knowledge of the local task and private data.
While \cite{zhu2021data, zhang2022fedzkt, zhang2022fine} transfer knowledge without any requirement of real data, all of them need high communication bandwidth due to the iterative exchange of models over hundreds of rounds, leading to high susceptibility to stealth attacks and, hence, privacy concerns. 

\section{Approach}
\subsection{Problem Statement}
Without loss of generality, we describe our method for the classification task in detail. 
Suppose that there are $K$ local nodes in a federated learning scenario, each privately holding a labeled dataset $ \{\domx'_k, \domy'_k \}$, consisting of the input image space $\domx' \in \mathbb{R}^{H \times W \times 3}$, and the label space $\domy' \in \{1, \dots, C \}$, where $C$ is the total number of classes. 

The proposed FedIOD includes two stages. First, with each private data $\{\domx'_k, \domy'_k \}$ we train the local model $T_k$ to complete. Note that the proposed FedIOD is agnostic to any neural network architecture. Hence, each local node can have its specialized architecture suited to the particular distribution of its local data. In the second stage, each locally trained model, $T_k$, will be frozen and only used as a teacher model in a one-way distillation paradigm. In contrast to \cite{gong2022federated, li2021practical} using carefully deliberated real data to transfer knowledge, we exploit ensemble knowledge in the input space $\mathcal{X}$ with a generator $G$ mapping from random noise $\domw$ to the input space $\mathcal{X}$. Taking such generated samples $x \sim \mathcal{X}$ as input, local models $T_k$ and the central task model $S$ on the server constitute a student-teacher knowledge transfer problem, with the teacher here being a group of local teachers. 
Let $\hat{\bm{z}}=S(x) $  and $\bm{z}_k=T_k(x)$ be the output logits of the central model and the $k$-th respectively ($\hat{\bm{z}}, \bm{z}_k \in \mathbb{R}^C$), the corresponding probability can be acquired with the following activation function:
\begin{equation}
\label{eq:softmax}
   p_\tau(\bm{z})= \left[\frac{e^{z^1/\tau}}{\sum_c{e^{ z^c/\tau}}}, \dots, \frac{e^{z^C/\tau}}{\sum_c{e^{ z^c/\tau}}}\right],
\end{equation}
where $\tau$ is a temperature parameter set to 1 by default. We abbreviate $p_\tau(\bm{z}_k)$ and $p_\tau(\hat{\bm{z}})$ as $\bm{q}_k = T_k (x; \tau) $ and $\hat{\bm{q}} = S(x; \tau)$, respectively. 

\subsection{Input Ensemble Distillation}
 To efficiently exploit the knowledge from local expertise, exploring the input space for the best fit of the global distribution is vital. We expect the optimal input to achieve (1) realism as a consensus achieved by all local nodes and (2) diversity to represent each local's unique knowledge under the heterogeneous federated learning scenarios.

\textbf{Consensual realism learning.}
Given the locally trained model $T_k$ as teachers and the central model $S$ as a student, we learn a generative model $G$ from randomly sampled noise $w$ to pseudo-data $x$, which will be the input for knowledge transfer. To guarantee the realism and practicality of $x$, we employ an additional discriminator $D_k$ residing at each local node to boost the generative model $G$ training.  $G$ is trained to approximate the global data distribution by fooling each local $D_k$. Following the typical training paradigm of GAN \cite{goodfellow2020generative, radford2015unsupervised}, we train $G$ and $D_k$ in a classical adversarial manner:
\begin{equation}
\label{eq:ganloss}
\begin{aligned}
    &\max_G \min_{D_k}L_\text{gan}^k(G, D_k) \\
    = &\max_G \min_{D_k} \fe_{x'_k \in \domx'_k} [\pi_k D_k(x_k')]  +  \fe_{w \in \domw} [1-D_k(G(w))],
\end{aligned}
\end{equation}
where $\pi_k =  \frac{|\domx'_k|}{\sum_{k'=1}^K |\domx'_{k'}|}$ is individual local weight and $|\domx'_k|$ indicates data size. In addition to this high-level realism, we expect $x$ to be realistic semantically, \ie, with semantic clarity according to the output of each locally trained model. Here, we assume that the input that confuses local models to produce ambiguous results will be less efficient in transferring knowledge. Hence, we expect each local model to produce confident predictions that the input $x$ tends to belong to one particular category. To force such semantic clarity, we maximize the confidence that $x$ belongs to one class. For each local node $k$, taking $\bm{q}_k$ as its corresponding probability, we minimize the Shannon entropy $H({\bm{q}})= - \sum_c \bm{q}^c \text{log} \bm{q}^c $, which can be reformulated as: 
\begin{equation}
\label{eq:confloss}
\begin{aligned}
    \min_{G} L_\text{conf}(G) &= \min_{G} \fe_{x \in \domx} [\sum_k{ \pi_k H (T_k(x; \tau))}] \\
    & =\min_{G}\fe_{w \in \domw} [\sum_k{\pi_k H (T_k(G(w); \tau))}]. \\
\end{aligned}
\end{equation}
\textbf{Per-local unique representation.}
The supervisions above ensure the realism of $x$, which are agreed upon by all local nodes. However, it can hardly transfer heterogeneous knowledge across local nodes. Our insight is that each local's expertise must be inconsistent, given the data heterogeneity in a federated learning scenario. Hence, the input must be diverse to generalize and transfer each local's unique knowledge.
To this point, we aim to generate $x$, which will tolerate local diversity, \wrt, input data on which local models produce divergent results. Specifically, we use Jensen-Shannon divergence to measure the dissimilarity of local probability outputs:
\begin{equation}
\label{eq:jsd}
    \begin{aligned}
        \text{JSD}(\bm{q}_1, \dots, \bm{q}_K) = H(\bar{\bm{q}}) - \sum_{k=1}^K \pi_k H(\bm{q}_k),
    \end{aligned}
\end{equation}
where $\bar{\bm{q}} = \sum_{k=1}^{K} \pi_k \bm{q}_k$ is the weighted ensemble of all locals.
We maximize such dissimilarity to encourage the level of local diversity, \wrt, unique local knowledge which has been exploited:
\begin{equation}
\label{eq:uniqueloss}
\begin{aligned}
    &\min_{G} L_\text{unique} (G) \\ 
    = & \min_{G} \fe_{w \in \domw} [-\text{JSD}(T_1(G(w); \tau), \dots, T_K(G(w); \tau)) ].\\
\end{aligned}
\end{equation}

\subsection{Output Ensemble Distillation}
Model distillation techniques typically optimize the student model by minimizing the KL divergence between the student model output $\hat{\bm{q}}$ and the teacher model output $\bar{\bm{q}}$ to bridge their performance gap:
\begin{equation}
\label{eq:kl}
\begin{aligned}
    \text{KL}(\bar{\bm{q}}|| \hat{\bm{q}}) = H(\bar{\bm{q}},\hat{\bm{q}}) - H(\hat{\bm{q}}),  \\
\end{aligned}
\end{equation}
where $H(\bar{\bm{q}}, \hat{\bm{q}})=-\sum_c \bar{\bm{q}}^c \log \hat{\bm{q}}^c$. Hinton \etal \cite{hinton2015distilling} has shown that minimizing Eq.~\ref{eq:kl} with a high $\tau$ (Eq.~\ref{eq:softmax}) is equivalent to minimizing the $\ell_2$ error between the logits of teacher and student, thereby relating cross-entropy minimization to fitting logits.  For multiple teachers, the conventional ensemble takes an average of all teachers' output probability as $\bar{\bm{q}}$.

However, under the FL scenario, it is not optimal to deploy such a local ensemble under the heterogeneous data distribution. This is mainly due to its inability to cope with the general setting when locally held data are not independent and identically distributed, \eg, they do not share precisely the same set of target classes. 
Let $P_{\domx'_k, \domy'_k}$ be the data distribution of the image and label over the $k$-th local data, and $P_{\domx', \domy'}$ be the global data distribution. Thus, we approximate the importance ratio of local prediction based on its training data distribution:
\begin{equation}
\label{eq:bias}
\begin{aligned}
    & \frac{P_{\domx'_k, \domy'_k}(y|x)}{P_{\domx', \domy'}(y|x)} = \frac{P_{\domy'_k}(y) P_{\domx'_k, \domy'_k}(x|y) P_{\domx'}(x)}{P_{\domy'}(y) P_{\domx', \domy'}(x|y) P_{\domx'_k}(x)} \\
    & \thickapprox  \frac{P_{\domy'_k}(y)}{P_{\domy'}(y)} \cdot \frac{P_{\domx'}(x)}{P_{\domx'_k}(x)} \thickapprox  \frac{P_{\domy'_k}(y)}{P_{\domy'}(y)} \cdot \frac{P_{\domx}(x)}{P_{\domx'_k}(x)} ,
\end{aligned}
\end{equation}
where we assume $P_{\domx'_k, \domy'_k}(x|y) \thickapprox  P_{\domx', \domy'}(x|y)$ as the local heterogeneity of this term is minor and ignorable compared to the heterogeneity in the image distribution $P_{\domx'}(x)$ and the label distribution $P_{\domy'_k}(y)$. And the global image distribution $\domx'$ is approximated with the generated input domain $\domx \thickapprox \domx'$. 

To consider this aspect, we introduce the weight of importance per class per input $\pi_k^c$ for each local node $k$ to reflect the data distribution with which its model was initially trained. Taking $x$ as input, we have the following. 

\begin{equation}
\label{eq:aggweight}
\begin{aligned}
\hat{\pi}_k^c(x) = \frac{\fe_{y'_k \in \domy'_k}|y'_k = c|}{{\fe_{ k \in \{1,\cdots,K\}, y'_k \in \domy'_k} |y'_k = c|}} \cdot \frac{D_k(x)}{\fe_{x'_k \in \domx'_k} D_k(x'_k)},
\end{aligned}
\end{equation}
where the first term corresponds to $\frac{P_{\domy'_k}(y)}{P_{\domy'}(y)}$ and can be acquired by statistics of local labels, \ie, the number of samples from class $c$ used to train the model at the local node $k$. The second term corresponds to $\frac{P_{\domx}(x)}{P_{\domx'_k}(x)}$ which can be approximated by the local discriminator's output on pseudo image $x$ and locally held image $x'_k$. We then normalize the importance weight between locals for each $c$: $\pi_k^c(x) = \hat{\pi}_k^c(x) / \sum_{k'=1}^{K} {\hat{\pi}_{k'}^c (x)}$.

\begin{algorithm}[t]
    \caption{FedIOD}
    \label{alg}
 \begin{algorithmic}
    \STATE {\bfseries Input:} Total number of local nodes $K$, locally held data $\{\domx'_k, \domy'_k\}$, local models $\{T_k\}$, central task model $S$, central generator $G$, auxiliary local discriminator $\{D_k\}$.
    \FOR{each local node $k=1,\cdots, K$} 
    \STATE Train $T_k$  with $(\domx'_k, \domy'_k)$ to complete
    \ENDFOR
    \FOR{each distillation step}
    \STATE {\color{gray} {$\Box$ Input distillation }}
    \STATE $w$ $\leftarrow$ randomly sampled from $\domw$ 
    \STATE $x \leftarrow G(w)$
    \FOR {$k =1 ,..., K$ }
    \STATE $\bm{z}_k$, $\bm{q}_k$ $\leftarrow T_k(x)$ 
    \STATE $x'_k$ $\leftarrow$ randomly sampled from $\domx'_k$
    \STATE $L_\text{gan}^k(G,D_k) \leftarrow D_k(x'_k), D_k(x)$ \hspace*{3em} $\triangleright$  Eq.~\ref{eq:ganloss}
    \STATE Update $D_k$ by descending its stochastic gradient  
    $\nabla_{D_k} L_\text{gan}$
    \ENDFOR
    \STATE $L_\text{conf}(G), L_\text{unique}(G) \leftarrow$ $\{\bm{q}_k\}$  \hspace*{5em}   $\triangleright$  Eq.~\ref{eq:confloss},\ref{eq:uniqueloss}
    \STATE {\color{gray} {$\Box$ Output distillation }}
    \STATE  $\hat{\bm{z}}$, $\hat{\bm{q}}$ $\leftarrow S(x)$ 
    \STATE $L_\text{mimic}(G, S) \leftarrow$  $\hat{\bm{z}}$, $\{\bm{z}_k\}$   \hspace*{6.7em}   $\triangleright$  Eq.~\ref{eq:mimicloss}
    \STATE {\color{gray} {$\Box$ Update }}
    \STATE Update $G$ by descending its stochastic gradient  
    $\nabla_{G} [ L_\text{conf} +  L_\text{unique} - L_\text{mimic} - \sum_{k=1}^K L_\text{gan}^k ]$
    \STATE Update $S$ by descending its stochastic gradient  
    $\nabla_{G} L_\text{mimic}$
    \ENDFOR
\end{algorithmic}
\end{algorithm}

Following the $\ell_2$ observation above of Hinton \etal \cite{hinton2015distilling}, we consider the case of $\tau \rightarrow \infty$ when deploying KL-divergence. Hence, it can be written as the $\ell_2$ error between central model logits $\hat{\bm{z}}$ and local aggregated $\bar{\bm{z}}$. Let $\bm{\pi}_k(x) =[\pi_k^1(x),\cdots,\pi_k^C(x)] \in [0,1]^C$ be the per-sample weight, and $\odot$ is Hadamard product, the local ensemble expertise is indicated as follows:
\begin{equation}
\label{eq:agglogits}
\begin{aligned}
    A(\bm{z}_1, \cdots, \bm{z}_K, x) = \sum_{k=1}^K \bm{\pi}_k(x) \odot \bm{z}_k,
\end{aligned}
\end{equation}
where the central model $S$ is optimized to mimic the local ensemble of expertise, while the generator $G$ is a critic to generate $x$ on which $S$ lags behind local experts. The motivation is that such challenging input will transfer the hard-to-mimic knowledge from local to central. Therefore, we tailor the input data on which the central model produces a result diverged from the local output. Using KL-divergence as a dissimilarity evaluation, we train $G$ and $S$ in an adversarial manner:
\begin{equation}
\label{eq:mimicloss}
\begin{aligned}
    &\max_G \min_S L_\text{mimic}(G, S)  
   = \\&\max_G \min_S 
   \fe_{w} 
   | S(G(w)) - A(T_1(G(w)), \cdots, T_K(G(w)))|^2,
\end{aligned}
\end{equation}
where $A(\cdot)$ is the aggregation function detailed in Eq.~\ref{eq:agglogits}.
To sum up, the overall loss function can be written as 
\begin{equation}
\label{eq:overallloss}
\begin{aligned}
   &\max_G \min_{D_k} L_\text{gan}^k(G, D_k) +\min_G [L_\text{conf}(G) + L_\text{unqiue}(G)] \\
   & +\max_G \min_S L_\text{mimic}(G, S).
\end{aligned}
\end{equation}
And the overall process is explained in Algorithm~\ref{alg}. 

\section{Experiments}
We provide comprehensive empirical studies with various heterogeneous FL settings on natural image classification and more privacy-sensitive medical tasks, including brain tumor segmentation and histopathological nuclei instance segmentation. 

\begin{table*}[]
\centering
\resizebox{\textwidth}{!}
{
\begin{tabular}{cc|c|c|cc|cc}
\toprule
\multicolumn{2}{c|}{\multirow{2}{*}{\textbf{Method}}} 
&Model- &Auxiliary  &
\multicolumn{2}{c|}{CIFAR-10} & 
\multicolumn{2}{c}{CIFAR-100} \\ 
& 
&agnostic & Prerequisite
&$\alpha=1$ &$\alpha=0.1$ &$\alpha=1$ &$\alpha=0.1$ \\
\midrule 
\multicolumn{2}{c|}{Standalone (mean $\pm$ std)}
&- &- &65.25$\pm$ 5.14 & 30.92$\pm$  11.17 &27.60$\pm$ 1.58 &16.99$\pm$ 2.46 
\\ \midrule
\parbox[t]{6mm}{\multirow{5}{*}{\rotatebox[origin=c]{90}{\shortstack[c]{Parameter- \\ based}}}} &FedAvg~\cite{mcmahan2017communication}
&\xmark &-
&
78.57$\pm$ 0.22 & 
68.37$\pm$ 0.50 
&
42.54$\pm$ 0.51 &
36.72$\pm$ 1.50 \\
&FedProx~\cite{li2018federated}
&\xmark &-
&
76.32$\pm$ 1.95 & 
68.65$\pm$ 0.77
&
42.94$\pm$ 1.23 & 
35.74$\pm$ 1.00 \\
&FedAvgM~\cite{hsu2019measuring}
&\xmark &-
&
77.79$\pm$ 1.22 & 
68.63$\pm$ 0.79 
&
42.83$\pm$ 0.36 & 
36.29$\pm$ 1.98 \\
&FedGEN~\cite{zhu2021data} 
&\xmark &task-relevant data 
&
80.31$\pm$ 0.97 & 
68.13$\pm$ 1.37  
&
45.97$\pm$ 0.23& 
35.97$\pm$ 0.31\\
&FedDF~\cite{lin2020ensemble}
&\xmark &task-relevant data 
&
\bf{80.69$\pm$ 0.43} & 
\bf{71.36$\pm$ 1.07}  
& 
\bf{47.43$\pm$ 0.45} & 
\bf{39.33$\pm$ 0.03} \\\midrule
\parbox[t]{6mm}{\multirow{3}{*}{\rotatebox[origin=c]{90}{\shortstack[c]{Distill- \\ based}}}} &FedMD~\cite{li2019fedmd}
&\cmark & task-relevant data 
&
80.37$\pm$ 0.37 & 
{69.23}$\pm$ 1.31 
&
\bf{45.83$\pm$ 0.58} & 
38.86$\pm$ 0.78 \\
&FedKD \cite{gong2022preserving}
&\cmark &task-relevant data 
&
{80.98}$\pm$ 0.11 & {65.46}$\pm$ 3.45 & {45.55}$\pm$ 0.38& {40.61}$\pm$ 2.54
\\
&FedIOD 
&\cmark &None &\bf{82.78$\pm$ 0.18} &\bf{70.08$\pm$ 0.37} &45.36$\pm$ 0.32 &\bf{41.88$\pm$ 0.16} \\
\bottomrule
\end{tabular}
}
\caption{Accuracy (\%) comparisons on the CIFAR-10 and CIFAR-100 datasets with ResNet-8 and $K$=20. ``Standalone'' indicates the performance of local models trained with individual private data. Several popular FL methods are compared with parameter-based and distillation-based FL prior arts, among which \cite{lin2020ensemble, zhu2021data} employ both parameter exchange and model output distillation.}
\label{tab:cifarcompare}
\end{table*}

\subsection{CIFAR-10/100 classification}
We use heterogeneous data splits with Dirichlet distribution following the prior art \cite{hsu2019measuring} for distributed local training sets. The value of $\alpha$ in the Dirichlet distribution controls the degree of non-IIDness: $\alpha \rightarrow \infty$ indicates an identical local data distribution, and a smaller $\alpha$ indicates a higher non-IIDness. We report average accuracy over three split seeds on the corresponding test set.

We conduct experiments following the typical FL setting \cite{lin2020ensemble} under $K$=20 and $\alpha$=1, 0.1 with ResNet-8. 
$w$ is randomly sampled with a dimension of 100, and $x=G(w)$ has a size of $32 \times 32$. We use a patch discriminator as $D_k$, of which the output is of size $8 \times 8$. 
The comparison in Table~\ref{tab:cifarcompare} shows that our method achieves superior or competitive results and a much stronger privacy guarantee.
Without the requirement of auxiliary data or prior knowledge of the local task, our method outperforms relevant-data-dependent distillation-based and parameter-based counterparts.
Moreover, our method demonstrates other benefits, including eliminating prerequisites of identical local model architecture or task-relevant real data.  

\subsection{Magnetic resonance image segmentation} 
We use the dataset from the 2018 Multimodal Brain Tumor Segmentation Challenge (BraTS 2018)~\cite{menze2014multimodal, bakas2018identifying}. 
Each subject was associated with voxel-level annotations of ``whole tumor", ``tumor core," and ``enhancing tumor." 
Following the experimental protocol of one prior art, \cite{chang2020synthetic}, we deploy 2D segmentation of the whole tumor on T2 images of HGG cases, among which 170 were for training and 40 for testing. The local data split also follows \cite{chang2020synthetic}.

\begin{table}[h]
\centering 
\resizebox{\columnwidth}{!}
{
\begin{tabular}
{c|cccc}
\toprule
&Dice(\%)$\uparrow$ &Sens.(\%)$\uparrow$ &Spec.(\%)$\uparrow$ &HD95(pixel)$\downarrow$ \\ \midrule
{Standalone} &\renewcommand\arraystretch{0.8} \begin{tabular}{@{}c@{}} 65.03 \\ \footnotesize{$\pm$3.31}\end{tabular} 
&\renewcommand\arraystretch{0.8} \begin{tabular}{@{}c@{}} 69.27  \\ \footnotesize{$\pm$4.72}\end{tabular} 
&\renewcommand\arraystretch{0.8} \begin{tabular}{@{}c@{}} 99.35 \\ \footnotesize{$\pm$0.15}\end{tabular} 
&\renewcommand\arraystretch{0.8} \begin{tabular}{@{}c@{}} 24.61 \\ \footnotesize{$\pm$3.62}\end{tabular}  \\\midrule
{Centralized} &74.85 &79.83 &99.55 &12.85 \\ \midrule
FedAvg &70.71 &67.31 &\bf{99.85} &{11.88}\\
AsynDGAN &70.43 &72.95 &99.57 &14.94\\
FedIOD &\bf{75.38} &\bf{79.47} &99.60 &\bf{11.76}\\
\bottomrule 
\end{tabular}}
\caption{Comparisons on the BraTS2018 dataset with $K$=10 under the same net with FedAvg and AsynDGAN. ``Centralized'': centralized training when all local data are collected together. 
}
\label{tab:brats}
\end{table}

We employ the same network structure of $G$, $D_k$, $S$, and the same data preprocessing as \cite{chang2020synthetic} for a fair comparison. Following its label condition $\domw$, we improve our $L_\text{gan}$ with additional perceptual loss \cite{johnson2016perceptual}.  
The Dice score, sensitivity (Sens.), specificity (Spec.), and Hausdorff distance (HD95) are used as evaluation metrics, where ``HD95'' represents 95\% quantile of the distances instead of the maximum.

Table \ref{tab:brats} compares our method with the prior art of distributed learning \cite{chang2020synthetic} and the classical parameter-based FedAvg method. Ours performs best segmentation on pixel-level overlap metrics (Dice and Sens.) and shape similarity metrics (HD95). 

\begin{table}
\centering 
\resizebox{\columnwidth}{!}
{
\begin{tabular}
{cc|cccc}
\toprule
& &Dice(\%)$\uparrow$ &Obj-Dice(\%)$\uparrow$  &AJI(\%)$\uparrow$ &HD95(pixel)$\downarrow$\\ \midrule
\parbox[t]{1mm}{\multirow{4}{*}{\rotatebox[origin=c]{90}{\shortstack[c]{{Standalone}}}}}
    &{breast} &77.92 &73.47 &53.64 &12.34\\
    &{liver} &79.16 &75.38 &55.63 &12.47 \\
    &{kidney} &74.99 &69.67 &50.99 &14.64\\
    &{prostate} &77.46 &73.74 &54.40 &15.59\\  \midrule
    \multicolumn{2}{c|}{FedAvg}  &78.12 &75.05 &55.56 &12.96\\
    \multicolumn{2}{c|}{AsynDGAN} &79.30 &72.73 &56.08 &14.49 \\
    \multicolumn{2}{c|}{FedIOD} &\bf{80.48} &\bf{77.03} &\bf{58.37} &\bf{11.22}\\
    \bottomrule 
\end{tabular}}
\caption{Comparisons on the TCGA dataset with four cross-organ local nodes. All methods use the same segmentation net provided by \cite{chang2020synthetic} for a fair comparison. }
\label{tab:nucleicrossorgan}
\end{table}

\subsection{Histopathological image segmentation} 
In real-world medical applications, the heterogeneity of data distributed among medical entities is not limited to the local size of the data or various subjects. Local data held by different clinical sites can be quite a domain variant, \eg, targeting different organs or collected with different infrastructures, which is relatively underexplored in contemporary FL methods. To this end, we evaluate our method in a cross-organ, cross-site setting where locally held data are from different organs and institutes. We experiment on nuclei instance segmentation task with pathological datasets, including TCGA~\cite{kumar2017dataset}, Cell17~\cite{vu2019methods} and TNBC~\cite{naylor2018segmentation}.  

\begin{table*}
\centering
\resizebox{\textwidth}{!}
{
\begin{tabular}{c|ccccc|cccc}
\toprule
&Test   &\multirow{2}{*}{Dice(\%)$\uparrow$} &\multirow{2}{*}{Obj-Dice(\%)$\uparrow$} &\multirow{2}{*}{{AJI(\%)$\uparrow$}} &\multirow{2}{*}{HD95(pixel)$\downarrow$} &\multicolumn{4}{c}{Average}\\
&Data & & & &  &Dice(\%)$\uparrow$ &{Obj-Dice(\%)$\uparrow$} &{AJI(\%)$\uparrow$} &HD95(pixel)$\downarrow$\\
\midrule
\multirow{3}{*}{FedAvg}  
&\cellcolor{gray0}{Cell17}  &\cellcolor{gray0}{68.74} &\cellcolor{gray0}{65.82} &\cellcolor{gray0}{39.37}  &\cellcolor{gray0}{24.15} &\multirow{3}{*}{63.42} &\multirow{3}{*}{64.00} &\multirow{3}{*}{37.29}  &\multirow{3}{*}{54.51} \\
&\cellcolor{gray1}{TCGA}  &\cellcolor{gray1}{77.57} &\cellcolor{gray1}{72.94} &\cellcolor{gray1}{50.03} &\cellcolor{gray1}{15.87} & & &  \\
&\cellcolor{gray2}{TNBC}  &\cellcolor{gray2}{43.95} &\cellcolor{gray2}{53.23} &\cellcolor{gray2}{22.48} &\cellcolor{gray2}{123.51} & & &  \\ \midrule
\multirow{3}{*}{AsynDGAN} 
&\cellcolor{gray0}{Cell17}  &\cellcolor{gray0}{79.82} &\cellcolor{gray0}{59.03} &\cellcolor{gray0}{34.64} &\cellcolor{gray0}{19.27} &\multirow{3}{*}{66.64}   &\multirow{3}{*}{61.15} &\multirow{3}{*}{34.21}  &\multirow{3}{*}{35.46} \\
&\cellcolor{gray1}{TCGA}  &\cellcolor{gray1}{52.29} &\cellcolor{gray1}{57.12} &\cellcolor{gray1}{26.03} &\cellcolor{gray1}{47.47} & & &  \\
&\cellcolor{gray2}{TNBC}  &\cellcolor{gray2}{67.80} &\cellcolor{gray2}{67.31} &\cellcolor{gray2}{41.96} &\cellcolor{gray2}{39.63} & & &  \\ \midrule
\multirow{3}{*}{FedIOD}  
&\cellcolor{gray0}{Cell17} &\cellcolor{gray0}{86.23} &\cellcolor{gray0}{68.03} &\cellcolor{gray0}{44.75} &\cellcolor{gray0}{7.01}  &\multirow{3}{*}{\bf{79.28}}  &\multirow{3}{*}{\bf{71.58}} &\multirow{3}{*}{\bf{49.52}} &\multirow{3}{*}{\bf{16.41}}  \\
&\cellcolor{gray1}{TCGA}  &\cellcolor{gray1}{76.59} &\cellcolor{gray1}{72.67} &\cellcolor{gray1}{53.04} &\cellcolor{gray1}{12.69} & & &  \\
&\cellcolor{gray2}{TNBC}  &\cellcolor{gray2}{75.01} &\cellcolor{gray2}{74.03} &\cellcolor{gray2}{50.76} &\cellcolor{gray2}{29.54} & & &  \\ 
\bottomrule
\end{tabular}}
\caption{Comparisons of cross-site cross-organ nuclei segmentation tasks with Cell17, TCGA, TNBC as distributed local data. For a fair comparison, all methods use the same U-Net architecture as the segmentation model and the same post-processing to convert the semantic prediction into instance segmentation results.
} 
\label{tab:nucleicrosssite}
\end{table*}

We cropped the images into patches of size $256 \times 256$ for training and inference. For metrics evaluation, the cropped patches are stitched back into the whole image with the original size. For $G$, $D_k$, and $S$,  we use the same model structure provided by \cite{chang2020synthetic} and the additional perceptual loss \cite{johnson2016perceptual} for $L_\text{gan}$.
We use object-level Dice \cite{chen2016dcan} and Aggregated Jaccard Index (AJI) \cite{vu2019methods} as metrics to evaluate the instance overlap or shape similarities for an individual object. Let $\bm{y}^i$ be the ground truth mask for the $i$-th instance of the total $n$ instances, and $\hat{\bm{y}}^j$ be the predicted mask for the $j$-th instance of the total $\hat{n}$ instances. 
$J(\bm{y}^i)= \text{argmax}_{\hat{\bm{y}}^j} {|\bm{y}^i \cap \hat{\bm{y}}^j|} / {|\bm{y}^i \cup \hat{\bm{y}}^j|}$ is the predicted instance that maximally overlaps $\bm{y}^i$, and $J(\hat{\bm{y}}^j)= \text{argmax}_{\bm{y}^i} {|\bm{y}^i \cap \hat{\bm{y}}^j|} / {|\bm{y}^i \cup \hat{\bm{y}}^j|}$ denotes the ground-truth instance that maximally overlaps $\hat{\bm{y}}^j$. 
For instance, for shape similarity, we use the Aggregated Jaccard Index (AJI): 
\begin{equation}
\label{eq:aji}
\begin{aligned}
    \text{AJI} (\bm{y}, \hat{\bm{y}})= \frac{\sum_{i=1}^{n} | \bm{y}^i \cap J(\bm{y}^i)| }{ \sum_{i=1}^{n} | \bm{y}^i \cup J(\bm{y}^i)| + \sum_{j \in \mathcal{J}} |\hat{\bm{y}}^j|},
\end{aligned}
\end{equation}
where $J(\bm{y}^i)$ is the predicted instance that has maximum overlap with $\bm{y}^i$ concerning the Jaccard index (sorted and nonrepeated). $\mathcal{J}$ is the set of predicted instances that have not been assigned to any ground-truth instance.

\begin{figure}[h]
\centering
\includegraphics[width=1.0\linewidth]{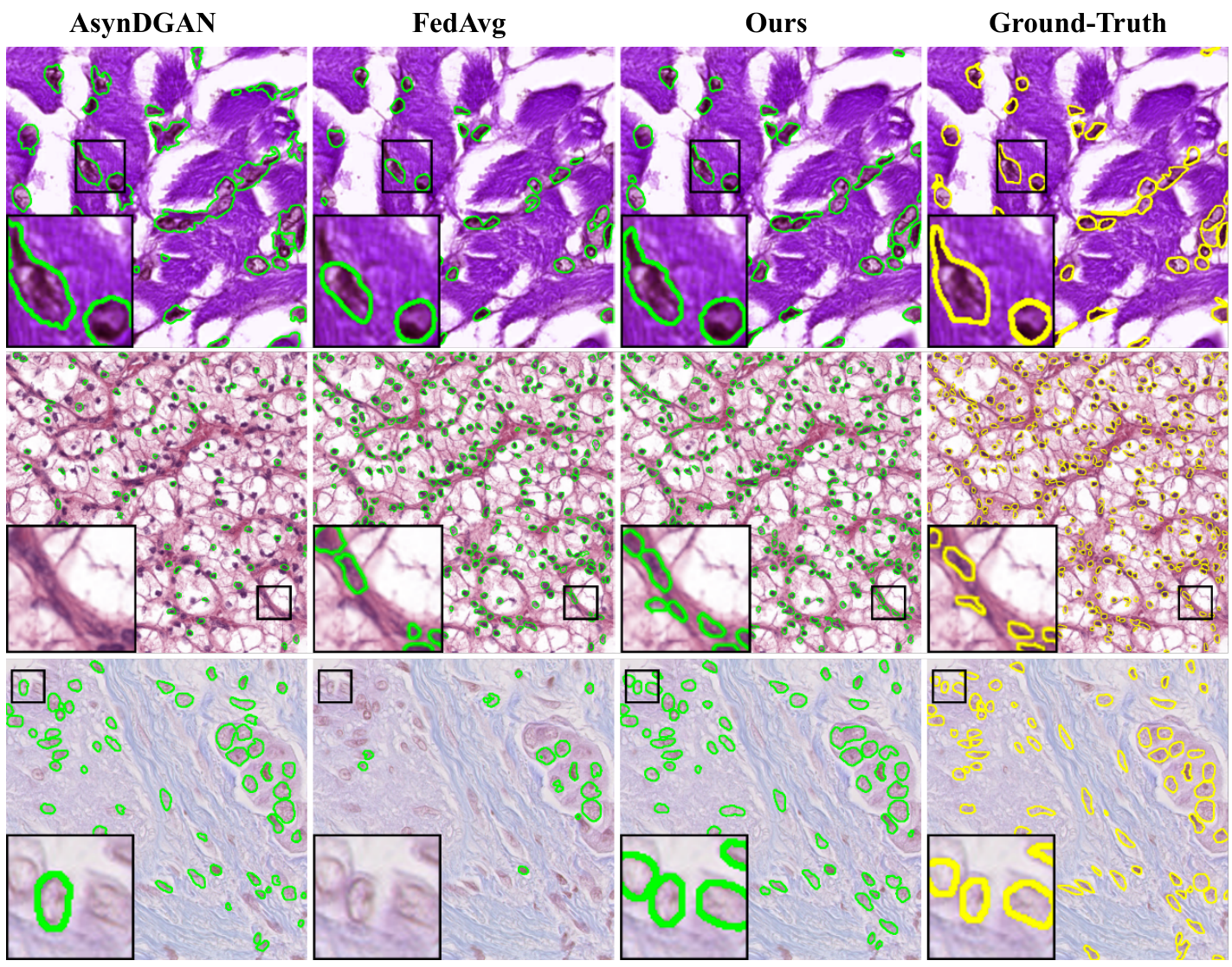}
\caption{Qualitative comparisons on cross-site cross-organ nuclei segmentation tasks. The three rows visualize instance contours on test images from Cell17, TCGA, and TNBC.} 
\label{fig3}
\end{figure}

\textbf{Cross-organ scenario.}
We first focus on cross-organ settings where each distributed local node holds only the data of one organ. Following \cite{chang2020synthetic}, from the TCGA dataset, we take 16 images of the breast, liver, kidney, and prostate for training and eight images of the same organs for testing. 
Table \ref{tab:nucleicrossorgan} shows the experimental results of this cross-organ setting and compares them with the baseline method \cite{chang2020synthetic} and the classical FedAvg. We can note that our method achieves the best results on semantic segmentation (Dice and Hausdorff) and instance segmentation (object-level Dice and AJI) metrics. 

\textbf{Cross-site cross-organ scenario.}
We also conduct experiments on more challenging settings with cross-site cross-organ datasets, where locally held data are from different organ nuclei datasets. 
Taking the training set of Cell17, TCGA, and TNBC as private data distributed over local nodes, we evaluate on the corresponding test sets. 
Table \ref{tab:nucleicrosssite} compares our method with two prior arts \cite{chang2020synthetic, mcmahan2017communication} on various segmentation metrics to evaluate semantic/instance level overlap and shape. 
Our proposed FedIOD outperforms the prior art on all these metrics for overlap and shape evaluation, 
demonstrating our efficacy in coping with heterogeneous FL scenarios. The qualitative comparisons shown in Figure \ref{fig3} also demonstrate the superiority of our method over its counterparts. 

\begin{table}[h]
    \centering
    \scalebox{1.0}{
    \begin{tabular}{c|ccccc}
    \hline
        \multicolumn{2}{c}{{Privacy budget $\varepsilon$ $\downarrow$}} & 3.5    & 6.0	 & 7.7	  & 10.0\\
        \hline
        \multirow{2}*{FedKD}
        & w/ DP $\uparrow$         &45.64	&56.08	&61.80	&70.90 \\
        & w/o DP $\uparrow$        &66.79   &79.30  &80.28  &81.55 \\\hline
        \multirow{2}*{FedIOD }
        & w/ DP $\uparrow$         &44.45	&58.96	&62.14	&73.58\\
        & w/o DP $\uparrow$        &74.31   &80.02  &82.03  &82.69 \\
    \hline
    \end{tabular}}
\noindent\caption{Compare FedIOD and FedKD in terms of accuracy (\%) on CIFAR10 ($K$=20, $\alpha$=1) under same privacy cost. 
} 
\label{tab:privacy-utility}
\end{table}

\section{Privacy Analysis}
\textbf{Comparison with data-dependent distillation-based FL.}
The significant difference between ours and typical FL based on distillation is that FedIOD generates data for knowledge distillation, while others rely on auxiliary real data. We adopt the differential privacy (DP) analysis in DP-CGAN \cite{torkzadehmahani2019dp} and GS-WGAN \cite{chen2020gs} to measure the privacy cost of the gradients used to train the generator. For a fair comparison, we apply PATE \cite{papernot2018scalable} on the local model output and then transfer them to the server to satisfy DP for both FedIOD and our counterpart FedKD~\cite{gong2022preserving}. Table \ref{tab:privacy-utility} compares FedIOD with FedKD in terms of accuracy under a series of rigid differential privacy protections ($\varepsilon <$10). 


\begin{figure}[h]
\centering
\includegraphics[width=0.95\linewidth]{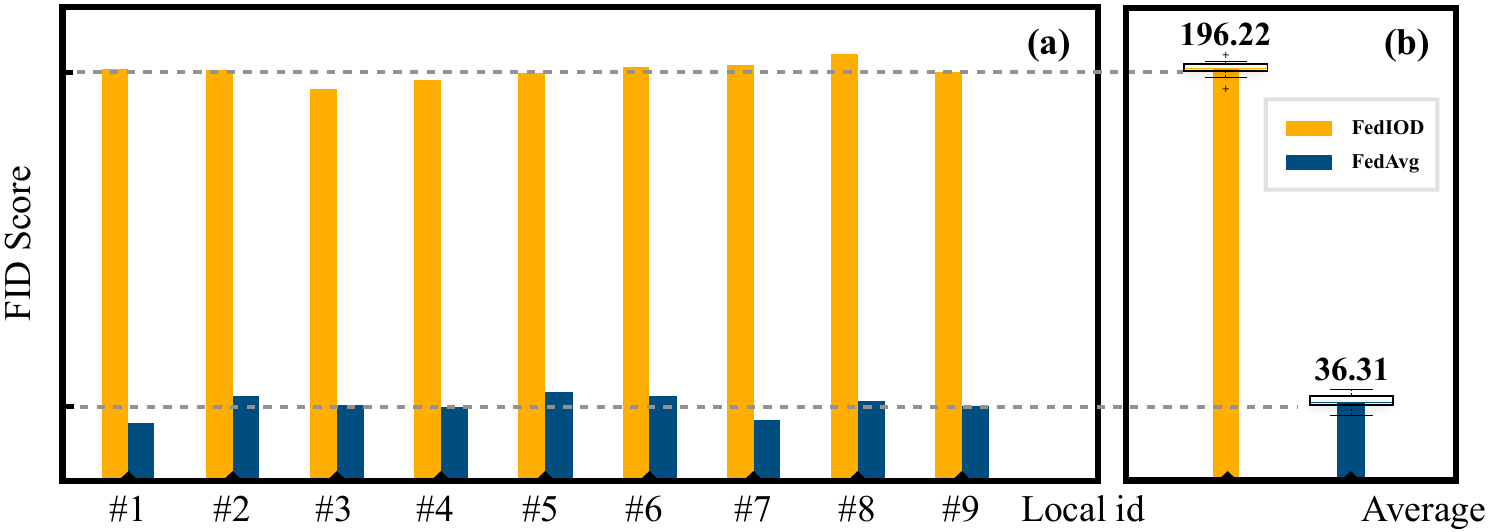}
\caption{ Comparison of FID scores between FedIOD and FedAvg on (a) 9 randomly selected local clients; and (b) average score under CIFAR10 ($K$=20, $\alpha$=1) FL setting. A larger FID indicates a stronger privacy guarantee.  
} 
\label{fig:fid}
\end{figure}


 \textbf{Comparison with parameter-based FL.} 
We use DLG \cite{zhu2019deep} as an attacker to recover private data using its iterative shared model parameters for parameter-based FL. We then measure the quality of the recovered data using Fréchet Inception Distance (FID). 
We assume a larger FID, \ie, a larger distance between the recovered data and private data, indicates a stronger privacy guarantee. For our method, we measure the FID between the synthetic images and the private images. The comparison in Figure \ref{fig:fid} shows that our method has a much higher FID, thus far more privacy protected than the FL parameter-sharing method such as FedAvg \cite{mcmahan2017communication}.  

\textit{Please refer to the ``Privacy Analysis'' section in the appendix for more details.}

\section{Conclusions}
In this work, we propose a novel federated learning framework, FedIOD, that protects local data privacy by distilling input and output to transfer knowledge from locals to the central server. To cope with the highly non-i.i.d. data distribution across local nodes, we learn the input on which each local achieves both consensual and unique results to represent individual heterogeneous expertise.  
We conducted extensive experiments with natural and medical images on classification and segmentation tasks in a variety of real, in-the-wild, heterogeneous FL settings.
All demonstrate the efficacy of FedIOD, showing its superior privacy-utility trade-off to others and significant flexibility in FL scenarios without any prerequisite of prior knowledge or auxiliary real data.

\section{Acknowledgment}
This research was supported in part by Zhejiang Provincial Natural Science Foundation of China under Grant No. D24F020011, Beijing Natural Science Foundation L223024, National Natural Science Foundation of China under Grant 62076016, the National Key Research and Development Program of China (Grant No. 2023YFC3300029) and “One Thousand Plan” projects in Jiangxi Province Jxsg2023102268 and a generous gift from Amazon.

\bibliography{aaai24}

\begin{thebibliography}{49}
\providecommand{\natexlab}[1]{#1}

\bibitem[{Asif, Tang, and Harrer(2019)}]{asif2019ensemble}
Asif, U.; Tang, J.; and Harrer, S. 2019.
\newblock Ensemble knowledge distillation for learning improved and efficient networks.
\newblock \emph{arXiv preprint arXiv:1909.08097}.

\bibitem[{Bakas et~al.(2018)Bakas, Reyes, Jakab, Bauer, Rempfler, Crimi, Shinohara, Berger, Ha, Rozycki et~al.}]{bakas2018identifying}
Bakas, S.; Reyes, M.; Jakab, A.; Bauer, S.; Rempfler, M.; Crimi, A.; Shinohara, R.~T.; Berger, C.; Ha, S.~M.; Rozycki, M.; et~al. 2018.
\newblock Identifying the best machine learning algorithms for brain tumor segmentation, progression assessment, and overall survival prediction in the BRATS challenge.
\newblock \emph{arXiv preprint arXiv:1811.02629}.

\bibitem[{Bakas(2020)}]{hdtd-5j88-20}
Bakas, S.~S. 2020.
\newblock Brats MICCAI Brain tumor dataset.

\bibitem[{Chang et~al.(2019)Chang, Shejwalkar, Shokri, and Houmansadr}]{chang2019cronus}
Chang, H.; Shejwalkar, V.; Shokri, R.; and Houmansadr, A. 2019.
\newblock Cronus: Robust and Heterogeneous Collaborative Learning with Black-Box Knowledge Transfer.
\newblock \emph{arXiv preprint arXiv:1912.11279}.

\bibitem[{Chang et~al.(2020)Chang, Qu, Zhang, Sabuncu, Chen, Zhang, and Metaxas}]{chang2020synthetic}
Chang, Q.; Qu, H.; Zhang, Y.; Sabuncu, M.; Chen, C.; Zhang, T.; and Metaxas, D.~N. 2020.
\newblock Synthetic learning: Learn from distributed asynchronized discriminator GAN without sharing medical image data.
\newblock In \emph{Proceedings of IEEE/CVF Conference on Computer Vision and Pattern Recognition}, 13856--13866.

\bibitem[{Chen, Orekondy, and Fritz(2020)}]{chen2020gs}
Chen, D.; Orekondy, T.; and Fritz, M. 2020.
\newblock Gs-wgan: A gradient-sanitized approach for learning differentially private generators.
\newblock \emph{Advances in Neural Information Processing Systems}, 33: 12673--12684.

\bibitem[{Chen et~al.(2016)Chen, Qi, Yu, and Heng}]{chen2016dcan}
Chen, H.; Qi, X.; Yu, L.; and Heng, P.-A. 2016.
\newblock DCAN: deep contour-aware networks for accurate gland segmentation.
\newblock In \emph{Proceedings of IEEE/CVF Conference on Computer Vision and Pattern Recognition}, 2487--2496.

\bibitem[{Chen et~al.(2019)Chen, Wang, Xu, Yang, Liu, Shi, Xu, Xu, and Tian}]{chen2019data}
Chen, H.; Wang, Y.; Xu, C.; Yang, Z.; Liu, C.; Shi, B.; Xu, C.; Xu, C.; and Tian, Q. 2019.
\newblock Data-free learning of student networks.
\newblock In \emph{Proceedings of IEEE/CVF International Conference on Computer Vision}, 3514--3522.

\bibitem[{Deng et~al.(2009)Deng, Dong, Socher, Li, Li, and Fei-Fei}]{deng2009imagenet}
Deng, J.; Dong, W.; Socher, R.; Li, L.-J.; Li, K.; and Fei-Fei, L. 2009.
\newblock Imagenet: A large-scale hierarchical image database.
\newblock In \emph{2009 IEEE conference on computer vision and pattern recognition}, 248--255. Ieee.

\bibitem[{Fang et~al.(2021)Fang, Bao, Song, Wang, Xie, Shen, and Song}]{fang2021mosaicking}
Fang, G.; Bao, Y.; Song, J.; Wang, X.; Xie, D.; Shen, C.; and Song, M. 2021.
\newblock Mosaicking to Distill: Knowledge Distillation from Out-of-Domain Data.
\newblock In \emph{Proceedings of Conference on Neural Information Processing Systems}.

\bibitem[{Fang et~al.(2019)Fang, Song, Shen, Wang, Chen, and Song}]{fang2019data}
Fang, G.; Song, J.; Shen, C.; Wang, X.; Chen, D.; and Song, M. 2019.
\newblock Data-free adversarial distillation.
\newblock \emph{arXiv preprint arXiv:1912.11006}.

\bibitem[{Geiping et~al.(2020)Geiping, Bauermeister, Dr{\"o}ge, and Moeller}]{geiping2020inverting}
Geiping, J.; Bauermeister, H.; Dr{\"o}ge, H.; and Moeller, M. 2020.
\newblock Inverting Gradients--How easy is it to break privacy in federated learning?
\newblock \emph{arXiv:2003.14053}.

\bibitem[{Gong et~al.(2022{\natexlab{a}})Gong, Sharma, Karanam, Wu, Chen, Doermann, and Innanje}]{gong2022preserving}
Gong, X.; Sharma, A.; Karanam, S.; Wu, Z.; Chen, T.; Doermann, D.; and Innanje, A. 2022{\natexlab{a}}.
\newblock Preserving Privacy in Federated Learning with Ensemble Cross-Domain Knowledge Distillation.
\newblock In \emph{Association for the Advancement of Artificial Intelligence}.

\bibitem[{Gong et~al.(2022{\natexlab{b}})Gong, Song, Vedula, Sharma, Zheng, Planche, Innanje, Chen, Yuan, Doermann, and Ziyan}]{gong2022federated}
Gong, X.; Song, L.; Vedula, R.; Sharma, A.; Zheng, M.; Planche, B.; Innanje, A.; Chen, T.; Yuan, J.; Doermann, D.; and Ziyan, W. 2022{\natexlab{b}}.
\newblock Federated Learning with Privacy-Preserving Ensemble Attention Distillation.
\newblock \emph{IEEE Transactions on Medical Imaging}.

\bibitem[{Goodfellow et~al.(2020)Goodfellow, Pouget-Abadie, Mirza, Xu, Warde-Farley, Ozair, Courville, and Bengio}]{goodfellow2020generative}
Goodfellow, I.; Pouget-Abadie, J.; Mirza, M.; Xu, B.; Warde-Farley, D.; Ozair, S.; Courville, A.; and Bengio, Y. 2020.
\newblock Generative adversarial networks.
\newblock \emph{Communications of the ACM}, 63(11): 139--144.

\bibitem[{Guo et~al.(2021)Guo, Wang, Zhou, Jiang, and Patel}]{Guo_2021_CVPR}
Guo, P.; Wang, P.; Zhou, J.; Jiang, S.; and Patel, V.~M. 2021.
\newblock Multi-Institutional Collaborations for Improving Deep Learning-Based Magnetic Resonance Image Reconstruction Using Federated Learning.
\newblock In \emph{Proceedings of IEEE/CVF Conference on Computer Vision and Pattern Recognition}.

\bibitem[{He et~al.(2016)He, Zhang, Ren, and Sun}]{he2016deep}
He, K.; Zhang, X.; Ren, S.; and Sun, J. 2016.
\newblock Deep residual learning for image recognition.
\newblock In \emph{Proceedings of the IEEE conference on computer vision and pattern recognition}, 770--778.

\bibitem[{Hinton, Vinyals, and Dean(2015)}]{hinton2015distilling}
Hinton, G.; Vinyals, O.; and Dean, J. 2015.
\newblock Distilling the knowledge in a neural network.
\newblock \emph{arXiv preprint arXiv:1503.02531}.

\bibitem[{Hsu, Qi, and Brown(2019)}]{hsu2019measuring}
Hsu, T.-M.~H.; Qi, H.; and Brown, M. 2019.
\newblock Measuring the effects of non-identical data distribution for federated visual classification.
\newblock \emph{arXiv preprint arXiv:1909.06335}.

\bibitem[{Hsu, Qi, and Brown(2020)}]{hsu2020federated}
Hsu, T.-M.~H.; Qi, H.; and Brown, M. 2020.
\newblock Federated Visual Classification with Real-World Data Distribution.
\newblock In \emph{Proceedings of European Conference on Computer Vision}.

\bibitem[{Isola et~al.(2017)Isola, Zhu, Zhou, and Efros}]{isola2017image}
Isola, P.; Zhu, J.-Y.; Zhou, T.; and Efros, A.~A. 2017.
\newblock Image-to-image translation with conditional adversarial networks.
\newblock In \emph{Proceedings of the IEEE conference on computer vision and pattern recognition}, 1125--1134.

\bibitem[{Jeong et~al.(2018)Jeong, Oh, Kim, Park, Bennis, and Kim}]{jeong2018communication}
Jeong, E.; Oh, S.; Kim, H.; Park, J.; Bennis, M.; and Kim, S.-L. 2018.
\newblock Communication-efficient on-device machine learning: Federated distillation and augmentation under non-iid private data.
\newblock \emph{arXiv preprint arXiv:1811.11479}.

\bibitem[{Johnson, Alahi, and Fei-Fei(2016)}]{johnson2016perceptual}
Johnson, J.; Alahi, A.; and Fei-Fei, L. 2016.
\newblock Perceptual losses for real-time style transfer and super-resolution.
\newblock In \emph{Proceedings of European conference on computer vision}, 694--711. Springer.

\bibitem[{Karimireddy et~al.(2020)Karimireddy, Kale, Mohri, Reddi, Stich, and Suresh}]{karimireddy2019scaffold}
Karimireddy, S.~P.; Kale, S.; Mohri, M.; Reddi, S.~J.; Stich, S.~U.; and Suresh, A.~T. 2020.
\newblock Scaffold: Stochastic controlled averaging for on-device federated learning.
\newblock In \emph{Proceedings of International Conference on Machine Learning}.

\bibitem[{Kumar et~al.(2017)Kumar, Verma, Sharma, Bhargava, Vahadane, and Sethi}]{kumar2017dataset}
Kumar, N.; Verma, R.; Sharma, S.; Bhargava, S.; Vahadane, A.; and Sethi, A. 2017.
\newblock A dataset and a technique for generalized nuclear segmentation for computational pathology.
\newblock \emph{IEEE Transactions on Medical Imaging}, 36(7): 1550--1560.

\bibitem[{Li and Wang(2019)}]{li2019fedmd}
Li, D.; and Wang, J. 2019.
\newblock Fedmd: Heterogenous federated learning via model distillation.
\newblock \emph{arXiv preprint arXiv:1910.03581}.

\bibitem[{Li, He, and Song(2021)}]{li2021practical}
Li, Q.; He, B.; and Song, D. 2021.
\newblock Practical one-shot federated learning for cross-silo setting.
\newblock In \emph{Proceedings of International Joint Conference on Artificial Intelligence}.

\bibitem[{Li et~al.(2018)Li, Sahu, Zaheer, Sanjabi, Talwalkar, and Smith}]{li2018federated}
Li, T.; Sahu, A.~K.; Zaheer, M.; Sanjabi, M.; Talwalkar, A.; and Smith, V. 2018.
\newblock Federated optimization in heterogeneous networks.
\newblock \emph{arXiv preprint arXiv:1812.06127}.

\bibitem[{Li et~al.(2020)Li, Sanjabi, Beirami, and Smith}]{li2019fair}
Li, T.; Sanjabi, M.; Beirami, A.; and Smith, V. 2020.
\newblock Fair resource allocation in federated learning.
\newblock In \emph{Proceedings of International Conference on Learning Representations}.

\bibitem[{Lin et~al.(2020)Lin, Kong, Stich, and Jaggi}]{lin2020ensemble}
Lin, T.; Kong, L.; Stich, S.~U.; and Jaggi, M. 2020.
\newblock Ensemble Distillation for Robust Model Fusion in Federated Learning.
\newblock In \emph{Proceedings of Conference on Neural Information Processing Systems}.

\bibitem[{McMahan et~al.(2017)McMahan, Moore, Ramage, Hampson, and y~Arcas}]{mcmahan2017communication}
McMahan, B.; Moore, E.; Ramage, D.; Hampson, S.; and y~Arcas, B.~A. 2017.
\newblock Communication-efficient learning of deep networks from decentralized data.
\newblock In \emph{Artificial Intelligence and Statistics}, 1273--1282. PMLR.

\bibitem[{Menze et~al.(2014)Menze, Jakab, Bauer, Kalpathy-Cramer, Farahani, Kirby, Burren, Porz, Slotboom, Wiest et~al.}]{menze2014multimodal}
Menze, B.~H.; Jakab, A.; Bauer, S.; Kalpathy-Cramer, J.; Farahani, K.; Kirby, J.; Burren, Y.; Porz, N.; Slotboom, J.; Wiest, R.; et~al. 2014.
\newblock The multimodal brain tumor image segmentation benchmark (BRATS).
\newblock \emph{IEEE Transactions on Medical Imaging}, 34(10): 1993--2024.

\bibitem[{Nayak et~al.(2019)Nayak, Mopuri, Shaj, Radhakrishnan, and Chakraborty}]{nayak2019zero}
Nayak, G.~K.; Mopuri, K.~R.; Shaj, V.; Radhakrishnan, V.~B.; and Chakraborty, A. 2019.
\newblock Zero-shot knowledge distillation in deep networks.
\newblock In \emph{Proceedings of International Conference on Machine Learning}, 4743--4751. PMLR.

\bibitem[{Naylor et~al.(2018)Naylor, La{\'e}, Reyal, and Walter}]{naylor2018segmentation}
Naylor, P.; La{\'e}, M.; Reyal, F.; and Walter, T. 2018.
\newblock Segmentation of nuclei in histopathology images by deep regression of the distance map.
\newblock \emph{IEEE Transactions on Medical Imaging}, 38(2): 448--459.

\bibitem[{Papernot et~al.(2018)Papernot, Song, Mironov, Raghunathan, Talwar, and Erlingsson}]{papernot2018scalable}
Papernot, N.; Song, S.; Mironov, I.; Raghunathan, A.; Talwar, K.; and Erlingsson, {\'U}. 2018.
\newblock Scalable private learning with pate.
\newblock \emph{arXiv preprint arXiv:1802.08908}.

\bibitem[{Radford, Metz, and Chintala(2015)}]{radford2015unsupervised}
Radford, A.; Metz, L.; and Chintala, S. 2015.
\newblock Unsupervised representation learning with deep convolutional generative adversarial networks.
\newblock \emph{arXiv preprint arXiv:1511.06434}.

\bibitem[{Ronneberger, Fischer, and Brox(2015)}]{ronneberger2015u}
Ronneberger, O.; Fischer, P.; and Brox, T. 2015.
\newblock U-net: Convolutional networks for biomedical image segmentation.
\newblock In \emph{Proceedings of International Conference on Medical Image Computing and Computer Assisted Intervention}.

\bibitem[{Szegedy et~al.(2017)Szegedy, Ioffe, Vanhoucke, and Alemi}]{szegedy2017inception}
Szegedy, C.; Ioffe, S.; Vanhoucke, V.; and Alemi, A.~A. 2017.
\newblock Inception-v4, inception-resnet and the impact of residual connections on learning.
\newblock In \emph{Thirty-first AAAI conference on artificial intelligence}.

\bibitem[{Torkzadehmahani, Kairouz, and Paten(2019)}]{torkzadehmahani2019dp}
Torkzadehmahani, R.; Kairouz, P.; and Paten, B. 2019.
\newblock Dp-cgan: Differentially private synthetic data and label generation.
\newblock In \emph{Proceedings of the IEEE/CVF Conference on Computer Vision and Pattern Recognition Workshops}, 0--0.

\bibitem[{Vu et~al.(2019)Vu, Graham, Kurc, To, Shaban, Qaiser, Koohbanani, Khurram, Kalpathy-Cramer, Zhao et~al.}]{vu2019methods}
Vu, Q.~D.; Graham, S.; Kurc, T.; To, M. N.~N.; Shaban, M.; Qaiser, T.; Koohbanani, N.~A.; Khurram, S.~A.; Kalpathy-Cramer, J.; Zhao, T.; et~al. 2019.
\newblock Methods for segmentation and classification of digital microscopy tissue images.
\newblock \emph{Frontiers in bioengineering and biotechnology}, 53.

\bibitem[{Wang et~al.(2020)Wang, Yurochkin, Sun, Papailiopoulos, and Khazaeni}]{wang2020federated}
Wang, H.; Yurochkin, M.; Sun, Y.; Papailiopoulos, D.; and Khazaeni, Y. 2020.
\newblock Federated learning with matched averaging.
\newblock In \emph{Proceedings of International Conference on Learning Representations}.

\bibitem[{Wu et~al.(2019)Wu, Zheng, Guo, and Lai}]{wu2019distilled}
Wu, A.; Zheng, W.-S.; Guo, X.; and Lai, J.-H. 2019.
\newblock Distilled person re-identification: Towards a more scalable system.
\newblock In \emph{Proceedings of IEEE/CVF Conference on Computer Vision and Pattern Recognition}, 1187--1196.

\bibitem[{Xiang, Ding, and Han(2020)}]{xiang2020learning}
Xiang, L.; Ding, G.; and Han, J. 2020.
\newblock Learning from multiple experts: Self-paced knowledge distillation for long-tailed classification.
\newblock In \emph{Proceedings of European Conference on Computer Vision}, 247--263. Springer.

\bibitem[{Yin et~al.(2020)Yin, Molchanov, Alvarez, Li, Mallya, Hoiem, Jha, and Kautz}]{yin2020dreaming}
Yin, H.; Molchanov, P.; Alvarez, J.~M.; Li, Z.; Mallya, A.; Hoiem, D.; Jha, N.~K.; and Kautz, J. 2020.
\newblock Dreaming to distill: Data-free knowledge transfer via deepinversion.
\newblock In \emph{Proceedings of IEEE/CVF Conference on Computer Vision and Pattern Recognition}, 8715--8724.

\bibitem[{Zbontar et~al.(2018)Zbontar, Knoll, Sriram, Murrell, Huang, Muckley, Defazio, Stern, Johnson, Bruno, Parente, Geras, Katsnelson, Chandarana, Zhang, Drozdzal, Romero, Rabbat, Vincent, Yakubova, Pinkerton, Wang, Owens, Zitnick, Recht, Sodickson, and Lui}]{zbontar2018fastMRI}
Zbontar, J.; Knoll, F.; Sriram, A.; Murrell, T.; Huang, Z.; Muckley, M.~J.; Defazio, A.; Stern, R.; Johnson, P.; Bruno, M.; Parente, M.; Geras, K.~J.; Katsnelson, J.; Chandarana, H.; Zhang, Z.; Drozdzal, M.; Romero, A.; Rabbat, M.; Vincent, P.; Yakubova, N.; Pinkerton, J.; Wang, D.; Owens, E.; Zitnick, C.~L.; Recht, M.~P.; Sodickson, D.~K.; and Lui, Y.~W. 2018.
\newblock {fastMRI}: An Open Dataset and Benchmarks for Accelerated {MRI}.
\newblock \emph{ArXiv 1811.08839}.

\bibitem[{Zhang et~al.(2022)Zhang, Shen, Ding, Tao, and Duan}]{zhang2022fine}
Zhang, L.; Shen, L.; Ding, L.; Tao, D.; and Duan, L.-Y. 2022.
\newblock Fine-tuning global model via data-free knowledge distillation for non-iid federated learning.
\newblock In \emph{Proceedings of IEEE/CVF Conference on Computer Vision and Pattern Recognition}, 10174--10183.

\bibitem[{Zhang, Wu, and Yuan(2022)}]{zhang2022fedzkt}
Zhang, L.; Wu, D.; and Yuan, X. 2022.
\newblock FedZKT: Zero-Shot Knowledge Transfer towards Resource-Constrained Federated Learning with Heterogeneous On-Device Models.
\newblock In \emph{2022 IEEE 42nd International Conference on Distributed Computing Systems}, 928--938. IEEE.

\bibitem[{Zhu, Liu, and Han(2019)}]{zhu2019deep}
Zhu, L.; Liu, Z.; and Han, S. 2019.
\newblock Deep leakage from gradients.
\newblock In \emph{Proceedings of Conference on Neural Information Processing Systems}, 14774--14784.

\bibitem[{Zhu, Hong, and Zhou(2021)}]{zhu2021data}
Zhu, Z.; Hong, J.; and Zhou, J. 2021.
\newblock Data-free knowledge distillation for heterogeneous federated learning.
\newblock In \emph{Proceedings of International Conference on Machine Learning}, 12878--12889. PMLR.

\end{thebibliography}


\newpage

\section*{\LARGE\textbf Appendix}

We provide materials supplementing the main manuscript, including the implementation details as well as some additional experiment results. 

\section{CIFAR-10/100 Classification}
The network of $G$ and $D_k$ adopt the same architecture of the generator and discriminator as that in \cite{fang2021mosaicking}. For a fair comparison, the network of $T_k$ and $S$ employ ResNet-8 following the prior art \cite{lin2020ensemble}. 

In the first stage, we train each local task model $T_k$ individually with SGD as optimizer and 0.0025 as learning rate. We adopt cross-entropy loss function and a batch size of 16 for 500 epochs. 
In the second stage, we update the generator $G$, discriminators $D_k$, and the central model $S$ simultaneously. We use Adam optimizer and Cosine Annealing decreasing the learning rate from 0.001 to 0 with a batch size of 64 for 300 epochs. We conduct an additional ablation study in table \ref{tab:cifarablation} to demonstrate the efficacy of each proposed module. 

In Table \ref{tab:cifars1}, we further show quantitative comparisons of the inception score (IS) of the synthetic transfer data. Typical inception score use InceptionNet \cite{szegedy2017inception} pretrained on ImageNet\cite{deng2009imagenet} to compute KL-divergence between the conditional and marginal probability distributions of the output. We adapt the inception score by inferring generated data with the locally trained model $T_k$ to evaluate its quality (each sample strongly classified as one class) and diversity (the overall probability of the generated data on each class of  $T_k$ tends to have even distribution). 

\begin{table}[b]
\begin{center}
\resizebox{\columnwidth}{!}
{
\begin{tabular}{cc|cccccccc}

\hline
$L_\text{gan}$  &$\pi_k$ &Avg &Eq.2 &Eq.2 &Eq.2 &Eq.2 &Eq.2 &Eq.2 &Eq.2\\ \hline
\multirow{2}{*}{$L_\text{mimic}$} &$\tau$ &\multirow{2}{*}{\xmark} &\multirow{2}{*}{\xmark} &1 & 1 &$\infty$ &$\infty$ &$\infty$ &$\infty$\\
&$\pi_k$ & &  &$*$ &Eq.8 &Eq.8 &Eq.8 &Eq.8 &Eq.8 \\\hline
\multicolumn{2}{c|}{$L_\text{conf}$} &\xmark &\xmark &\xmark &\xmark &\xmark &\cmark &\xmark &\cmark\\\hline
\multicolumn{2}{c|}{$L_\text{unique}$} &\xmark &\xmark &\xmark &\xmark &\xmark &\xmark &\cmark &\cmark\\ \hline
\multicolumn{2}{c|}{Acc. (\%)} &57.03 &65.05 &59.88 &65.85 &68.49 &69.21 &69.67 &70.36 \\
\hline
\end{tabular}}
\end{center}
\caption{Ablation study on CIFAR-10 with ResNet-8, $K$=20 and $\alpha$=0.1. For the training of $L_\text{gan}$, we compare our weighting scheme (Eq. 2) with the typical average ensemble.  For the ensemble scheme of $L_\text{mimic}$, we compare our per-sample, per-class importance weighting (Eq. 8) with $*$ which represents the weighting scheme commonly used in other FL methods \cite{lin2020ensemble, hsu2020federated}. To compare $\tau$, we only list the result with a typical value $\tau$=1~\cite{hinton2015distilling}.}
\label{tab:cifarablation}
\end{table}

\begin{table}[t]
\begin{center}
\resizebox{\columnwidth}{!}
{
\begin{tabular}{c|ccccc}
\hline
$L_\text{gan}(G)$  &\cmark &\cmark &\cmark &\cmark &\cmark \\\hline
{$L_\text{mimic}(G)$} &\xmark  &$*$ &Eq.8  &Eq.8  &Eq.8 \\\hline
{$L_\text{conf}(G)$} &\xmark &\xmark &\xmark  &\cmark &\cmark\\\hline
{$L_\text{unique}(G)$} &\xmark &\xmark &\xmark  &\xmark &\cmark\\\hline
{Inception Score $\uparrow$} &2.40 &2.90 &2.95 &2.64 &3.57 \\
{Adapted Inception Score $\uparrow$} &2.30 &2.19 &2.35 &2.74 &2.82 \\
\hline
\end{tabular}}

\end{center}
\caption{Ablation study on the fidelity of the generated data, with $K$=20 and $\alpha$=0.1 on CIFAR-10. For the ensemble scheme of $L_\text{mimic}$, we compare our per-sample per-class importance weighting (Eq. 8) with the weighting scheme commonly used in other FL methods (represented with $*$) \cite{lin2020ensemble, hsu2020federated}. We compare both typical inception score and adapted inception score  evaluated by each locally trained model $T_k$ (taking average of all local models).}
\label{tab:cifars1}
\end{table}

\begin{figure}[h]
\centering
\includegraphics[width=\linewidth]{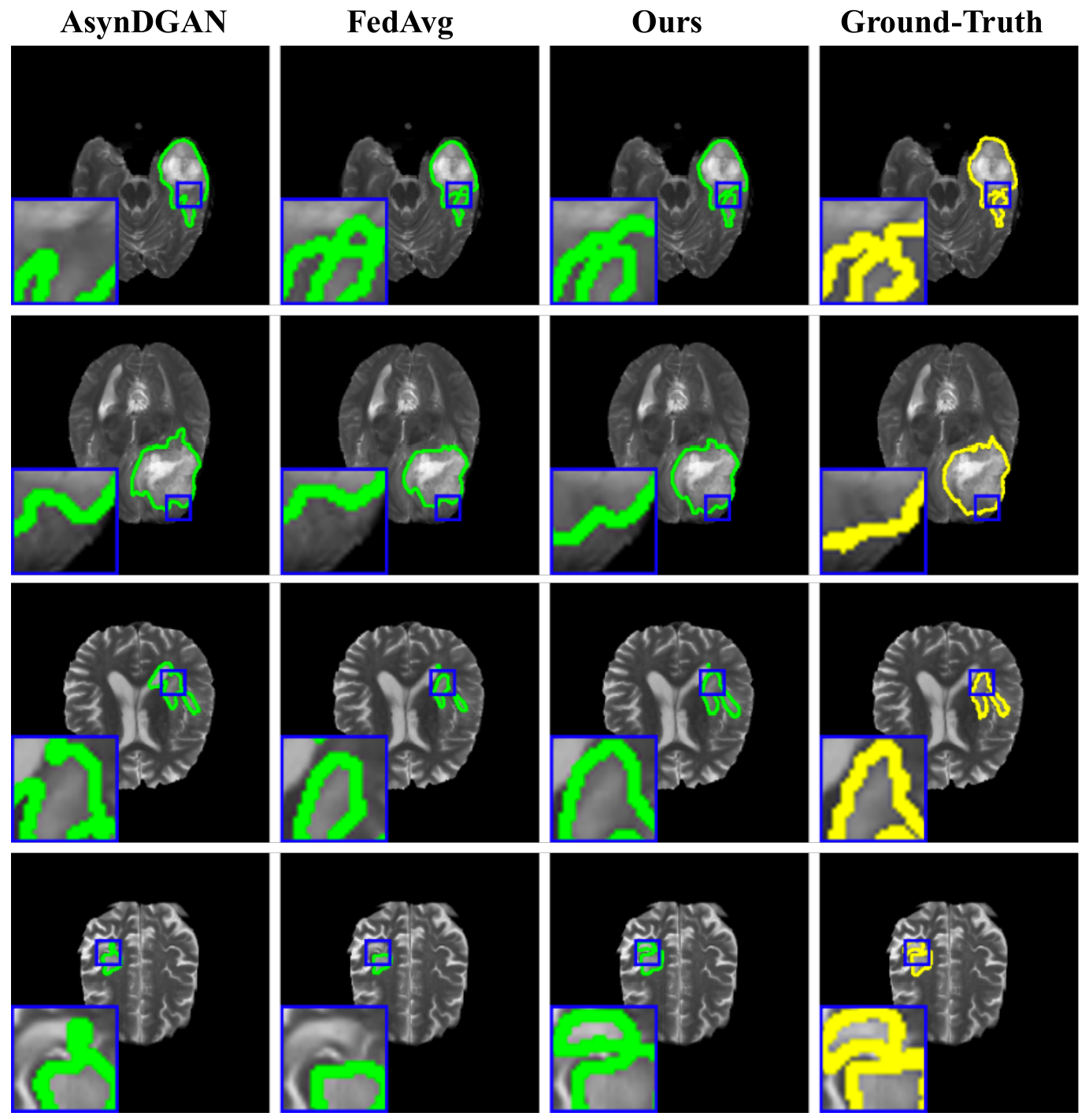}
\caption{Visualization of testing results on BraTS2018 dataset with $K$=10. We compare ours with AsynDGAN~\cite{chang2020synthetic} and FedAvg~\cite{mcmahan2017communication}.  We highlight the contours extracted from each method's segmentation prediction as well as the ground-truth. The zoomed part is shown at the left-bottom of each image and demonstrates that our method achieves much closer prediction to the ground-truth.
} 
\label{fig:s1}
\end{figure}

\section{Magnetic resonance image segmentation}
The 2018 Multimodal Brain Tumor Segmentation Challenge (BraTS 2018)~\cite{menze2014multimodal, bakas2018identifying} contains multi-parametric preoperative magnetic resonance imaging scans of 285 subjects with brain tumors, including 210 high-grade glioma (HGG) and 75 low-grade gliomas (LGG) subjects. 
Each subject was associated with voxel-level annotations of “whole tumor”, “tumor core”, and “enhancing tumor”. Each subject was scanned under the T1-weighted, T1-weighted with contrast enhancement, T2-weighted, and T2 fluid-attenuated inversion recovery (T2-FLAIR) modalities. Following the experimental protocol of one prior art \cite{chang2020synthetic}, we deploy 2D segmentation of the whole tumor on T2 images of HGG cases, among which 170 were for training and 40 for testing. The local data split also follows \cite{chang2020synthetic}: we first sort the training cases with tumor size and then divide the training set into ten subsets distributed to 10 local nodes. Overall there are 11,057 slices as training images across all local nodes and 2,616 slices as testing images. Following \cite{chang2020synthetic}, the network structure of $G$ employs a 9-block ResNet \cite{he2016deep}, and each discriminator $D_k$ employs the same structure as the patch discriminator in \cite{isola2017image}. The segmentation net for $T_k$ and $S$ follow the same U-Net \cite{ronneberger2015u} structure as that in \cite{chang2020synthetic}. 

In the first stage of local training,
we employ Adam optimizer and a learning rate of 0.002 to train each local model $T_k$ using cross-entropy loss and dice loss. The batch size is 16 and the total number of training epochs is 50. During training, we crop and resize the image to $224 \times 224$ following the same procedure as that in \cite{chang2020synthetic}. In the second stage of distillation, we use the label condition with size $240 \times 240$ following \cite{chang2020synthetic} and improve our $L_\text{gan}$ with additional perceptual loss \cite{johnson2016perceptual}.
We adopt the Adam optimizer with a learning rate of 0.0002 and batch size of 2 for 400 epochs. We randomly crop the generated image $x$ to $224 \times 224$ and randomly rotate and flip images as data augmentation during distillation. 

The Dice score, sensitivity, specificity, and Hausdorff distance are used as evaluation metrics. Taking $\bm{y}, \hat{\bm{y}} \in \{0,1 \}^{H \times W}$ as the ground-truth mask and the segmentation prediction, respectively, Dice evaluates the overlap between the two: $\text{Dice}(\bm{y}, \hat{\bm{y}}) = {2|\bm{y} \cap \hat{\bm{y}} |} / {(|\bm{y}|+|\hat{\bm{y}}|)}$. Sensitivity represents the true positive rate: $\text{Sens}(\bm{y}, \hat{\bm{y}}) = {|\bm{y} \cap \hat{\bm{y}} |} / {|\bm{y}|}$, and specificity represents the true negative rate: $\text{Spec}(\bm{y}, \hat{\bm{y}}) = {|(1-\bm{y}) \cap (1-\hat{\bm{y}}) |} / {|1-\bm{y}|}$. The Hausdorff distance evaluates the shape similarity: 
\begin{equation}
\label{eq:hausdorff}
    \text{HD}(\bm{y},\hat{\bm{y}}) = \max \{\sup_{\bm{u} \in \partial{\bm{y}}} \inf_{\bm{\hat{u}} \in \partial{\bm{\hat{y}}}}  |\bm{u}-\hat{\bm{u}}|, \sup_{\bm{\hat{u}}\in \partial{\bm{\hat{y}}}} \inf_{\bm{u} \in \partial{\bm{y}}}    |\bm{u}-\hat{\bm{u}}| \},
\end{equation}
where $\partial$ indicates boundary extraction and returns boundary position sets. ``HD95'' represents 95\% quantile of the distances instead of the maximum.

In Figure \ref{fig:s1}, we show qualitative results on the segmentation performance of central segmentation net $S$. We can note that our method outperforms the other two counterparts \cite{mcmahan2017communication, chang2020synthetic} on tumor shape segmentation with much more closer prediction to the ground-truth.

\section{Histopathological image segmentation}
\subsection{Datasets}
\textbf{TCGA:} 
The TCGA dataset~\cite{kumar2017dataset}  was captured from the Cancer Genome Atlas  archive and used in MICCAI 2018 multi-organ segmentation challenge (MoNuSeg). The training set consists of 30 images and around 22,000 nuclei instance annotations, while the test set includes 14 images with additional 7000 nuclei boundary annotations. The images are with $1000$ $\times$ $1000$ pixels and captured at 40$\times$  magnification on hematoxylin and eosin (H\&E) stained tissue. These images show highly varying properties from 18 hospitals and seven organs (breast, liver, kidney, prostate, bladder, colon, and stomach).  

\noindent \textbf{Cell17:}  
The MICCAI 2017 Digital Pathology Challenge dataset~\cite{vu2019methods} (Cell17)  consists of 64 H\&E stained histology images. Both the training and testing sets contain 32 images from four different diseases: glioblastoma multiforme (GBM), lower-grade glioma (LGG) tumors, head, and neck squamous cell carcinoma (HNSCC), and non-small cell lung cancer (NSCLC). The image sizes are either $500 \times 500$ or $600 \times 600$ at $20\times$ or $40\times$ magnification.

\noindent \textbf{TNBC:}  
The Triple Negative Breast Cancer (TNBC)~\cite{naylor2018segmentation} dataset consists of 50 annotated $512 \times 512$ images at $40\times$ magnification. The images are sampled from 11 patients at the Curie Institute, with three to eight images for each patient. Overall there are  4022 annotated cell instances. The image data includes low cellularity regions, which can be stromal areas or adipose tissue, and high cellularity areas consisting of invasive breast carcinoma cells.

\begin{table*}
\centering
\scalebox{0.9}
{
\begin{tabular}{cccccc|cccc}
\toprule
Train
&Test   &\multirow{2}{*}{Dice(\%)$\uparrow$} &\multirow{2}{*}{Obj-Dice(\%)$\uparrow$} &\multirow{2}{*}{{AJI(\%)$\uparrow$}} &\multirow{2}{*}{HD95(pixel)$\downarrow$} &\multicolumn{4}{c}{Average}\\
Data &Data & & & &  &Dice(\%)$\uparrow$ &{Obj-Dice(\%)$\uparrow$} &{AJI(\%)$\uparrow$} &HD95(pixel)$\downarrow$\\
\midrule
\multirow{3}{*}{Cell17}  &\cellcolor{gray0}{Cell17} &\cellcolor{gray0}{85.90} &\cellcolor{gray0}{67.24} &\cellcolor{gray0}{44.26} &\cellcolor{gray0}{8.21} &\multirow{3}{*}{59.37}  &\multirow{3}{*}{50.59} &\multirow{3}{*}{27.42}  &\multirow{3}{*}{32.89}\\  
&\cellcolor{gray1}{TCGA} &\cellcolor{gray1}{43.03} &\cellcolor{gray1}{33.83} &\cellcolor{gray1}{11.75} &\cellcolor{gray1}{34.75} & & &  \\
&\cellcolor{gray2}{TNBC} &\cellcolor{gray2}{49.18} &\cellcolor{gray2}{50.70} &\cellcolor{gray2}{26.26} &\cellcolor{gray2}{55.71} & & &  \\ \midrule
\multirow{3}{*}{TCGA}   &\cellcolor{gray0}{Cell17} &\cellcolor{gray0}{55.10} &\cellcolor{gray0}{48.25} &\cellcolor{gray0}{29.42} &\cellcolor{gray0}{31.94} &\multirow{3}{*}{49.20}  &\multirow{3}{*}{48.41} &\multirow{3}{*}{32.94} &\multirow{3}{*}{57.07}  \\ 
&\cellcolor{gray1}{TCGA} &\cellcolor{gray1}{75.75} &\cellcolor{gray1}{71.58} &\cellcolor{gray1}{51.91} &\cellcolor{gray1}{13.14} & & &  \\
&\cellcolor{gray2}{TNBC}  &\cellcolor{gray2}{16.77} &\cellcolor{gray2}{25.42} &\cellcolor{gray2}{7.50} &\cellcolor{gray2}{126.14} & & &  \\ \midrule
\multirow{3}{*}{TNBC} &\cellcolor{gray0}{Cell17} &\cellcolor{gray0}{70.42} &\cellcolor{gray0}{56.79} &\cellcolor{gray0}{36.59} &\cellcolor{gray0}{24.33} &\multirow{3}{*}{61.36}  &\multirow{3}{*}{52.65} &\multirow{3}{*}{33.35} &\multirow{3}{*}{27.78} \\
&\cellcolor{gray1}{TCGA} &\cellcolor{gray1}{35.59} &\cellcolor{gray1}{24.32} &\cellcolor{gray1}{5.65} &\cellcolor{gray1}{35.31} & & &  \\
&\cellcolor{gray2}{TNBC} &\cellcolor{gray2}{78.08} &\cellcolor{gray2}{76.84} &\cellcolor{gray2}{57.81} &\cellcolor{gray2}{23.72} & & &  \\ 
\bottomrule
\end{tabular}}
\caption{The performance of locally trained models under the cross-site cross-organ nuclei segmentation setting with Cell17~\cite{vu2019methods}, TCGA~\cite{kumar2017dataset}, TNBC~\cite{naylor2018segmentation} as distributed local data. 
} 
\label{tab:nucleis2}
\end{table*}

\begin{figure}[h]
\centering
\includegraphics[width=\linewidth]{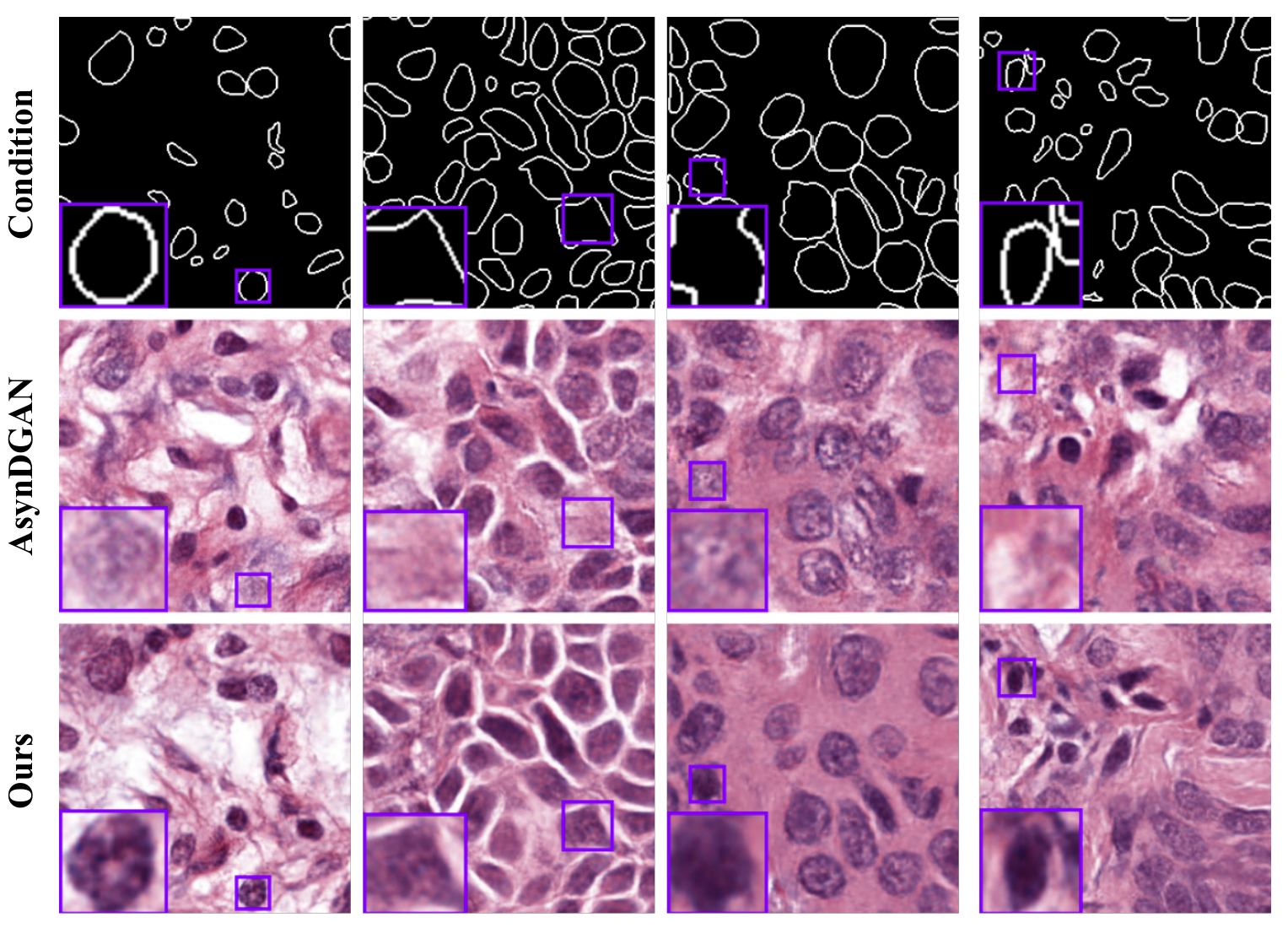}
\caption{Visualization of synthetic data on cross-organ TCGA dataset with $K$=4. We compare ours with AsynDGAN~\cite{chang2020synthetic}.  We zoom the instance region at the left-bottom of each image where our method succeeds to generate corresponding nuclei instances while the counterpart fails.
} 
\label{fig:s2}
\end{figure}

\subsection{Implementation details and results}
Following \cite{chang2020synthetic}, the network structure of $G$ employs a 9-block ResNet \cite{he2016deep}, and each discriminator $D_k$ employs the same structure as the patch discriminator in \cite{isola2017image}. The segmentation net for $T_k$ and $S$ follow the same U-Net \cite{ronneberger2015u} structure as that in \cite{chang2020synthetic}. 

In the first stage of local training, each local model $T_k$ is trained with Adam optimizer and a constant learning rate of $2.5\times10^{-4}$. The batch size is 4 and the overall number of training epochs is 100. We employ weighted cross-entropy loss where the foreground and the contour region are given more weight than the background region.
The data augmentation during training includes random rotation, random cropping ($256 \times 256$), and random flip both horizontally and vertically. 
In the second stage of distillation, we use the same label as condition with size $256 \times 256$ following \cite{chang2020synthetic} and improve our $L_\text{gan}$ with additional perceptual loss \cite{johnson2016perceptual}. $G$ and $D_k$ are pretrained for 100 epochs and then trained together with $S$ for another 300 epochs. We use Adam optimizer with a learning rate of 0.0001 and batch size of 8.

Figure \ref{fig:s2} shows the visualization of synthetic images under the cross-organ experiment setting. From the comparisons of the highlighted region, we can note that our synthetic data used for knowledge transfer achieves much better qualitative results (clear instance generation given the instance contour) over the counterpart \cite{chang2020synthetic}. Table \ref{tab:nucleis2} shows the performance of locally trained models under the cross-site cross-organ nuclei segmentation (corresponding to Table 4 in the main manuscript).

\begin{figure*}
\centering
\includegraphics[width=\linewidth]{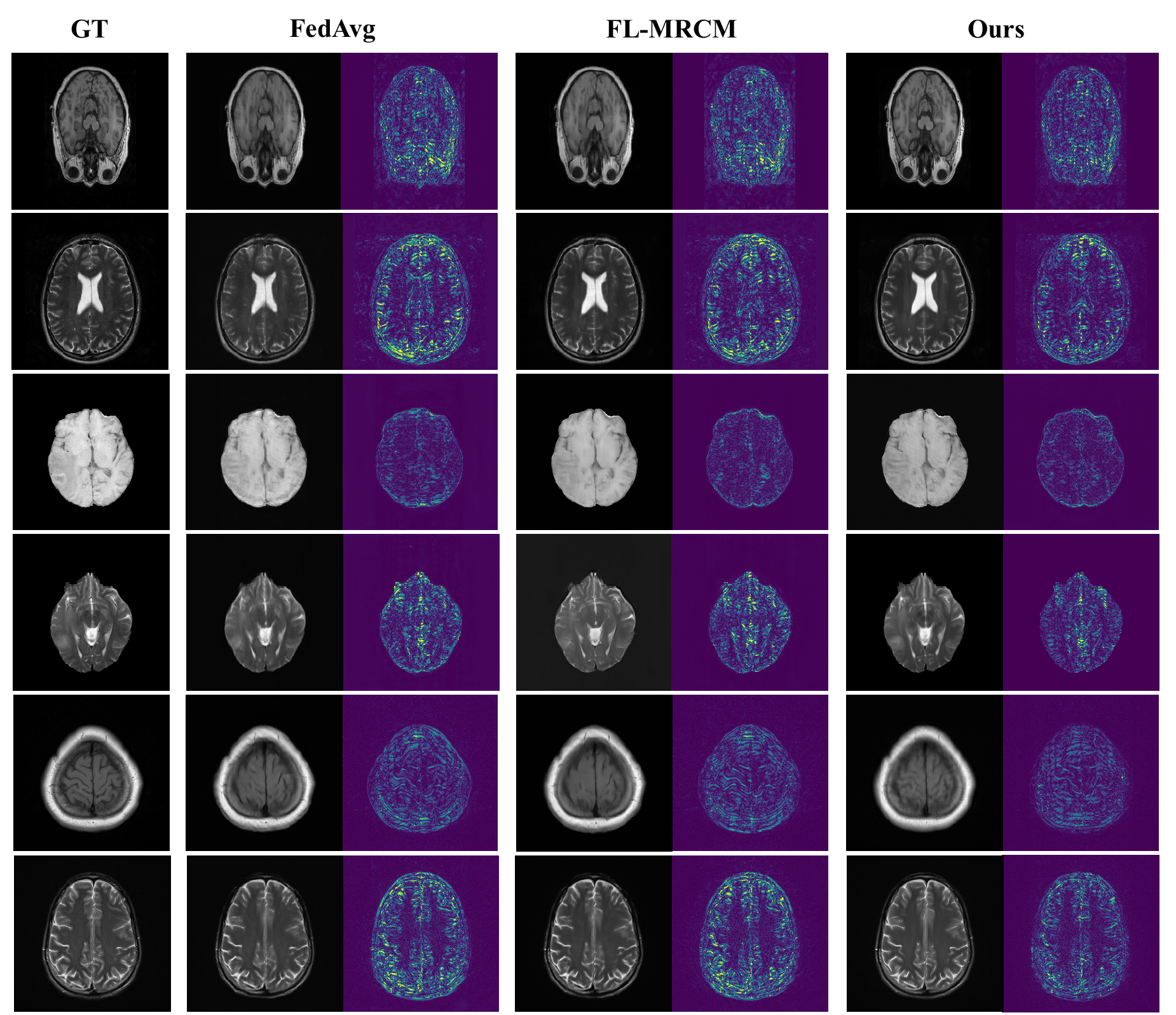}
\caption{Visualization of MRI image reconstruction with IXI, BraTS2020 and fastMRI as locally held data. We compare ours with two other FL methods: FedAvg and FL-MRCM. Each FL method trains with T1/T2-
weighted IXI, BraTS2020, fastMRI as local data and tests on T1  IXI, T2  IXI, T1 BraTS2020, T2 BraTS2020, T1 fastMRI, T2 fastMRI test set respectively. The second column of each sub-figure is the error map (absolute difference) between the reconstructed images and the ground truth (GT).
}

\label{fig:s3}
\end{figure*}

\section{Brain MRI reconstruction}
The proposed FedIOD framework can be used for other tasks, \eg, magnetic resonance image reconstruction. Following the prior-art experiment protocol \cite{Guo_2021_CVPR, gong2022federated}, we use fastMRI \cite{zbontar2018fastMRI}, IXI \footnote{\url{https://brain-development.org/}}, BraTS\cite{hdtd-5j88-20} as private data distributed across local nodes and evaluate the corresponding test sets. 




\begin{table}
\centering
\resizebox{\columnwidth}{!}
{
\begin{tabular}{cccc|cc|cc}
\toprule
&\multirow{2}{*}{NO shared } &\multirow{3}{*}{Prerequisite}
&Test &\multicolumn{2}{c|}{{{T1-weighted}}} &\multicolumn{2}{c}{{{T2-weighted}}}  \\
\cmidrule{5-8}
&Param. &  &{Data} &{SSIM$\uparrow$} &{PSNR$\uparrow$} &{SSIM$\uparrow$} &{PSNR$\uparrow$}  \\
\midrule
\multirow{3}{*}{FedAvg}
&\multirow{3}{*}{\xmark} 
&{Identical} &\cellcolor{gray0}B &\cellcolor{gray0}{0.9317} &\cellcolor{gray0}{34.91} 
&\cellcolor{gray0}{0.9173} &\cellcolor{gray0}{30.68}  
\\
& &{model}  &\cellcolor{gray1}F  &\cellcolor{gray1}{0.8803} &\cellcolor{gray1}30.48 &\cellcolor{gray1}{0.8782} &\cellcolor{gray1}{30.09} \\
& &{structure} &\cellcolor{gray2}I &\cellcolor{gray2}{0.9232} &\cellcolor{gray2}{32.31} &\cellcolor{gray2}0.8597 &\cellcolor{gray2}29.89 \\
\midrule

\multirow{3}{*}{FL-MRCM}
&\multirow{3}{*}{\xmark} 
&{Identical} &\cellcolor{gray0}B &\cellcolor{gray0}{0.9577} &\cellcolor{gray0}{36.88} 
&\cellcolor{gray0}{0.9308} &\cellcolor{gray0}{34.28}  
\\
& &{model} &\cellcolor{gray1}F &\cellcolor{gray1}0.9023 &\cellcolor{gray1}{33.63} &\cellcolor{gray1}0.8974 &\cellcolor{gray1}31.24 \\
& &{structure} &\cellcolor{gray2}I &\cellcolor{gray2}{0.9362} &\cellcolor{gray2}{33.29} &\cellcolor{gray2}0.8778 &\cellcolor{gray2}30.44  \\
\midrule \midrule

\multirow{3}{*}{FedAD}
&\multirow{3}{*}{\cmark} 
&Auxiliary &\cellcolor{gray0}B &\cellcolor{gray0}0.9111 &\cellcolor{gray0}34.55 
&\cellcolor{gray0}0.9199 &\cellcolor{gray0}34.06  
\\
& &task-relevant  &\cellcolor{gray1}F &\cellcolor{gray1}{0.9182} &\cellcolor{gray1}33.37 &\cellcolor{gray1}{0.9374} &\cellcolor{gray1}{32.76}\\
& &image  &\cellcolor{gray2}I &\cellcolor{gray2}0.9173 &\cellcolor{gray2}31.72  &\cellcolor{gray2}{0.9058} &\cellcolor{gray2}{30.93}\\
\midrule
\multirow{3}{*}{FedIOD} 
&\multirow{3}{*}{\cmark} 
&Auxiliary &\cellcolor{gray0}B 
&\cellcolor{gray0}{0.9326}  &\cellcolor{gray0}{36.08} 
&\cellcolor{gray0}0.9064 &\cellcolor{gray0}34.39  
\\
& &task-relevant  &\cellcolor{gray1}F &\cellcolor{gray1}0.8725 &\cellcolor{gray1}30.53 &\cellcolor{gray1}0.9057 &\cellcolor{gray1}30.45  \\
& &label  &\cellcolor{gray2}I &\cellcolor{gray2}{0.9198} &\cellcolor{gray2}{32.15} &\cellcolor{gray2}0.8650 &\cellcolor{gray2}29.69 \\
\bottomrule
\end{tabular}}
\caption{Results on cross-domain MRI image reconstruction with fastMRI, BraTS2020, and IXI as locally held data (abbreviated as F, B, I respectively). We compare SSIM and PSNR with parameter-based methods, FedAvg and FL-MRCM,  as well as distillation-based prior art FedAD.
}
\label{tab:mriin}
\end{table}

We use the same preprocessing and U-Net \cite{ronneberger2015u} architecture for the reconstruction networks as \cite{Guo_2021_CVPR, gong2022federated}. 
Compared to distillation-based methods, we achieve competitive results 
with far fewer prerequisites: our counterpart \cite{gong2022federated} relies on additional brain MRI images with the same modalities for distillation. At the same time, ours only utilizes the contour of the foreground of the brain as a condition of $G$, demonstrating much more relaxation and flexibility.  In addition, our method achieves comparable SSIM and PSNR with parameter-based methods while simultaneously demonstrating other benefits, including protecting privacy by not sharing local parameters.
\subsection{Datasets}
\textbf{fastMRI} \cite{zbontar2018fastMRI}:  For fastMRI T1-weighted images, we use 2,583 subjects for training and 860 for testing. For T2-weighted images, 2,874 subjects are used for training and 958 for testing. Each subject consists of approximately 15 axial cross-sectional images of brain tissues.

\noindent \textbf{BraTS2020} \cite{hdtd-5j88-20}: BraTS2020 consists of 494 subjects for both  T1 and T2-weighted modalities. There are 369 subjects for training and 125 subjects for testing. Each subject includes approximately 120 axial cross-sectional images of brain tissues for both modalities.

\noindent \textbf{IXI}: 
IXI T1-weighted images include 436, 55, and 90 subjects for training, validation, and testing, respectively. 
For the T2-weighted modality there are 434, 55, and 89 subjects for training, validation, and testing.  Each subject includes approximately 150 and 130 axial cross-sectional images of brain tissues for T1 and T2-weighted respectively.

\subsection{Implementation details and results}
The architecture of $G$ and $D_k$ are the same as those used in brain tumor image segmentation. And the reconstruction network $T_k$ and $S$ follow the same U-Net architecture as that in \cite{Guo_2021_CVPR}. 

For local training, we train each $T_k$ with Adam optimizer and a constant learning rate of 0.0001 for 20 epochs following \cite{gong2022federated}. 
For the second stage of distillation, we update the networks with Adam optimizer and a constant learning rate of 0.0001 for 100 epochs. In Figure \ref{fig:s3} we show qualitative results of the reconstructed images as well as the comparisons with two other FL methods  \cite{mcmahan2017communication, Guo_2021_CVPR}.

\section{Privacy Analysis}
\textbf{Comparison with data-dependent distillation-based FL.}
The major difference between ours and typical FL based on distillation is that FedIOD generates data for knowledge distillation, while others rely on auxiliary real data. Although eliminating such a prerequisite of real data, the gradients backpropagated to train the generator might raise security concerns. To this point, we adopt the differential privacy (DP) analysis in DP-CGAN \cite{torkzadehmahani2019dp} and GS-WGAN \cite{chen2020gs} to measure the privacy cost of the gradients used to train the generator. By clipping and adding Gaussian noise to these gradients, it satisfies $(\varepsilon, \delta)$-differential privacy: it allows a small probability ($\delta = 10^{-5}$) for the privacy budget $\varepsilon$. For a fair comparison, we apply PATE \cite{papernot2018scalable} on the local model output and then transfer them to the server to satisfy DP for both FedIOD and our counterpart FedKD~\cite{gong2022preserving}. Table \ref{tab:privacy-utility} compares FedIOD with FedKD in terms of accuracy under a series of rigid differential privacy protections ($\varepsilon <$10). We can see that our method (a) eliminates the requirements of prior knowledge of the local task and task-relevant public data during federated distillation; (b) and at the same time achieves superior or equivalent performance to FedKD under the same privacy cost.



\begin{figure}[b]
\centering
\includegraphics[width=\linewidth]{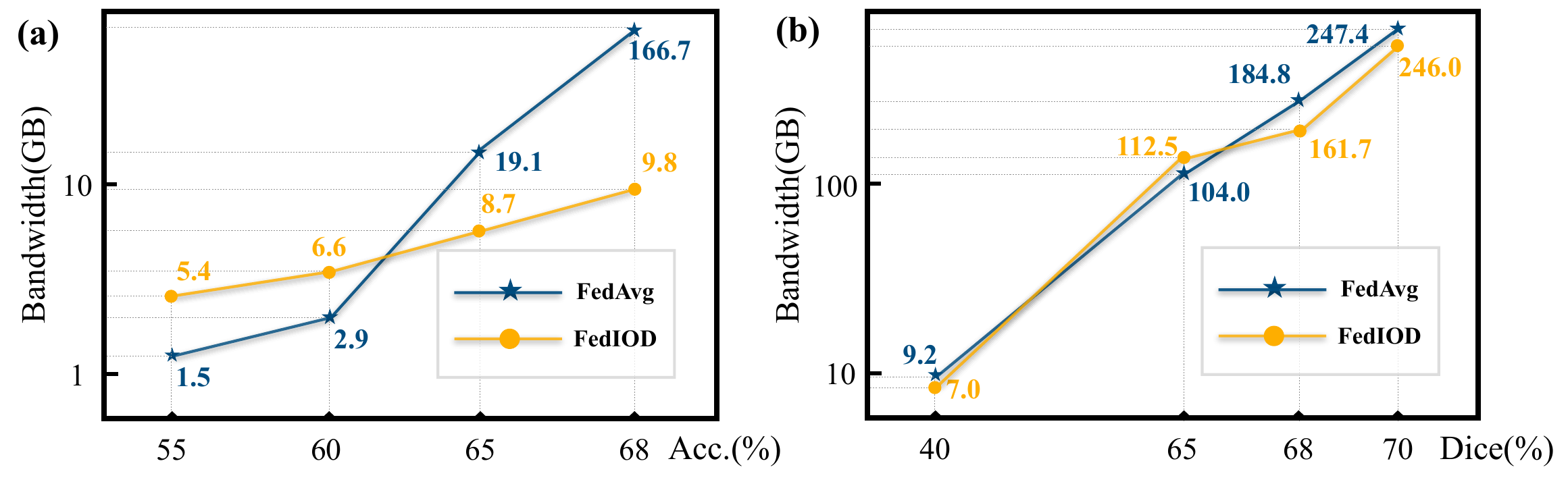}
\caption{ Comparisons of communication cost for (a) CIFAR10 ($K$=20, $\alpha$=0.1) classification; and (b) BraTS2018 segmentation to reach certain performance.
} 
\label{fig:bandwidth}
\end{figure}

 \textbf{Comparison with parameter-based FL.} 
Sharing parameters makes it vulnerable to white-box attacks \cite{chang2019cronus, zhu2019deep, geiping2020inverting}, while our distillation-based method only has black-box attack risks. Although it is intuitive that distillation-based FL is more secure than parameter-based FL, the synthetic images used in distillation-based FedIOD may raise privacy concerns. We use the similarity between synthetic images and privately held local data as a quantization of privacy leakage. 
For parameter-based FL, we use DLG \cite{zhu2019deep} as an attacker to recover private data using its iterative shared model parameters. We then measure the quality of the recovered data using Fréchet Inception Distance (FID). We assume a larger FID, \ie, a larger distance between the recovered data and private data, indicates a stronger privacy guarantee. For our method, we measure the FID between the synthetic images and the private images. The comparison in Figure \ref{fig:fid} shows that our method has a much higher FID, thus far more privacy protected than the FL parameter-sharing method such as FedAvg \cite{mcmahan2017communication}.  
In particular, our proposed method outperforms the parameter sharing method \cite{mcmahan2017communication} and simultaneously provides a much more secure privacy guarantee. Figure~\ref{fig:bandwidth} shows that our method costs less or equivalent communication bandwidth compared to the parameter-based art \cite{mcmahan2017communication} .

\end{document}